\documentclass[11pt]{article}
\usepackage[letterpaper]{geometry}

\usepackage{hyperref}
\usepackage{graphicx}
\usepackage{multirow}
\usepackage{amssymb}
\usepackage{amsmath}
\usepackage[vlined]{algorithm2e}
\usepackage{algpseudocode}
\usepackage[bottom]{footmisc}
\usepackage[plain]{algorithm}
\usepackage{xcolor}

\begin{document}

\title{\Large %Prediction of Depression Treatment %Response \\
%using Motif Discovery\\
Motif Discovery Framework for Psychiatric EEG Data Classification
%\relatedversion
}
\author{Melanija Kraljevska*
\and Kate\v{r}ina Hlav\'a\v{c}kov\'{a}-Schindler* 
\and Lukas Miklautz* 
\and Claudia Plant \thanks{Research Group Data Mining and Machine Learning, Faculty of Computer Science, University of Vienna,
 W\"ahringerstrasse 29, 1090 Vienna, Austria and ds:UniVie, University of Vienna, Austria}
}

\date{}

\maketitle

% Copyright Statement
% When submitting your final paper to a SIAM proceedings, it is requested that you include
% the appropriate copyright in the footer of the paper.  The copyright added should be
% consistent with the copyright selected on the copyright form submitted with the paper.
% Please note that "20XX" should be changed to the year of the meeting.

% Default Copyright Statement
%\fancyfoot[R]{\scriptsize{Copyright \textcopyright\ 2023 by SIAM\\
%Unauthorized reproduction of this article is prohibited}}

% Depending on which copyright you agree to when you sign the copyright form, the copyright
% can be changed to one of the following after commenting out the default copyright statement
% above.

%\fancyfoot[R]{\scriptsize{Copyright \textcopyright\ 20XX\\
%Copyright for this paper is retained by authors}}

%\fancyfoot[R]{\scriptsize{Copyright \textcopyright\ 20XX\\
%Copyright retained by principal author's organization}}

%\pagenumbering{arabic}
%\setcounter{page}{1}%Leave this line commented out.

%\vspace{1cm}

\begin{abstract}  
In current medical practice, patients undergoing depression treatment must wait four to six weeks before a clinician
can assess medication response due to the delayed noticeable effects of antidepressants.  
Identification of a treatment response at  any earlier stage is of great importance, since it can reduce the emotional and
economic burden connected with the treatment.
We approach the prediction of a patient response to a treatment as a
classification problem, by utilizing the dynamic properties of EEG recordings on the 7{th} day of the treatment. 
We present a novel framework that  applies motif discovery to extract meaningful features from EEG data  distinguishing between depression treatment responders and non-responders.  We applied our framework also to classification tasks in other psychiatric EEG datasets, namely to patients with 
symptoms of schizophrenia, pediatric patients with intractable seizures, and Alzheimer disease and dementia.
We achieved high classification precision in all data sets. The results demonstrate that the dynamic
properties of the EEGs may support clinicians in decision making both in diagnosis and  in the prediction depression treatment response as early as on the 7th day of the treatment.
 To our best knowledge, our work is the first one using motifs  in the depression diagnostics in general. 
\end{abstract}

%\begin{figure}
%\vspace{14pc}
%\caption{This is a figure 1.1.}
%\end{figure}

\section{Introduction}

\begin{figure}
\centering
\includegraphics[width=12cm]{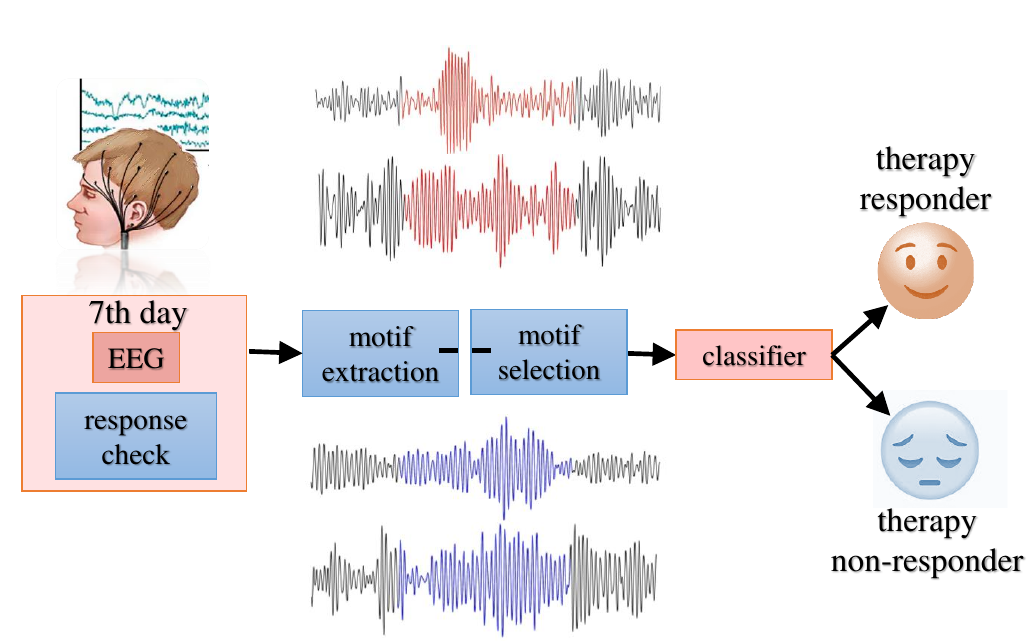}
%\vspace{-0.2cm}
\caption{Overview on our treatment prediction framework}
\label{teaser}
\end{figure}

Depression represents a common mental disorder that affects people globally. When
diagnosing depression, medical practitioners check for the presence of the debilitating
disease called Major Depressive Disorder (MDD).
%It is characterized by sad and depressive
%moods, reduced interests, cognitive dysfunction, and physical symptoms such as appetite
%or sleep disturbances \cite{otte2016major}. 
%MDD differs from regular mood changes and
%negative emotions from everyday life, as it can have an impact on various aspects of one’s life and society in general.
%MDD
%affects women twice as often as men and it occurs in approximately one out of every six
%adults at some point in their lives.
It is estimated that 3.8$\%$ of the population suffers from depression
%which amounts to approximately 280 million people worldwide 
\cite{whodepression}.
%Depression affects 5$\%$ of adults and 5.7$\%$
%with age above 60. Additionally, the risk of suicide in people with MDD is about 20 times
%that of the general population \cite{whodepression}.
The treatments of depression include psychological treatment and  antidepressant
medication in more serious cases.
%however, due to insufficient resources, antidepressants are used more frequently
%than psychological interventions.
However, the response rate of antidepressants in average, i.e.,
the percentage of cases of an improvement of depression symptoms, is in the
range of 42-53 $\%$ \cite{taliaz2021optimizing}. A patient undergoing treatment needs to wait 4 to 6 weeks
before getting checked by a clinician whether they are responding to the medication, as the
antidepressants take time to produce noticeable effects of alleviated depressive symptoms
 \cite{gautam2017clinical}. 
 To assess the effectiveness of the antidepressant treatment and monitor changes
over time, 
%it is a common practice for healthcare providers to use
the Montgomery-Åsberg
Depression Rating Scale (MADRS) \cite{madrs1979new} is used in practice. The MADRS questionnaire measures the
severity of depression in the individual; a higher score indicates worse symptoms. If the
MADRS score obtained after 4--6 weeks of treatment shows no improvement in the
patient’s symptoms, then the treatment needs to be changed or adjusted.
%Although antidepressive treatments have been effective in treating depression for some patients, there are several limitations and downsides associated with the common clinical practice.
 Due to the latency of the drug effect, in case of non-responsiveness to the
medication, the patient can endure a considerable amount of distress and waste time on
ineffective treatments. Electroencephalography (EEG)  is a cost-effective clinical tool.
To reduce the emotional and economic burden of a patient, in this work we propose a  motif-based  framework (see a schematic overview in Figure \ref{teaser}) which based on the EEG of the 7th day of treatment predicts whether a patient will   respond  to the medication   or not.

We approach the prediction of patient response to treatment as a classification problem.
%by utilizing the dynamic properties of EEG recordings of depression patients. 
In order to build a classifier for earlier prediction of responsiveness of treatment, features extracted from the EEGs of the 7th day  of the treatment are used. For training, we use as class labels the  MADRS score of the patient in the 28th day of the treatment. We perform the so-called motif discovery in EEGs of depressive patients and utilize the identified motifs in the feature engineering step, with the aim of differentiating between responsive and non-responsive patients. 
%motifs

The goal of motif discovery is to identify frequent, unknown patterns in a time series, without the pre-existing knowledge about their location and shape \cite{torkamani2017survey}.
% OLD: The goal of motif discovery is to identify frequent, unknown patterns in a time series. In our domain, there is no pre-existing knowledge about their location and shape of the patterns that differentiate responders from non-responders, therefore we follow an unsupervised approach \cite{torkamani2017survey}.
Motifs are defined in the literature as short time series that represent reoccurring patterns, frequent trends, or approximately repeated sequences  \cite{motifdef12003probabilistic}\cite{motifdef22004discovering}\cite{motifdef22007detecting}.
%Motifs can capture characteristic temporal changes and allow for the identification and representation of relevant patterns at different scales and lengths over time.
In the case of EEG, motifs can correspond to specific brain activities or states, e.g., different stages of being asleep can be differentiated using motifs detected in EEG sleep data \cite{kohlmorgen2000identification}. %Detecting such reoccurring patterns in EEGs could aid in understanding different brain behaviours between patient groups, as they can be easily visualized and inspected by domain experts.
% challenges with motif discovery 
There are several challenges when discovering motifs; as motifs represent patterns that are similar to each other, the similarity measure needs to account for possible noise and different scales, amplitudes, and variability of the patterns throughout the signal. Moreover, detecting motifs with unknown lengths requires the algorithm to be flexible enough to handle a wide range of possible lengths, which increases computational complexity in the case of high-dimensional or large-scale data \cite{torkamani2017survey}. 
%Additionally, discovering ill-known or warped motifs can be a challenge, as the algorithm must be able to identify patterns that may not conform to a strict or well-defined shape. 
%In the case of EEG with multivariate time series, we want to discover the so-called consensus motifs that represent conserved patterns that occur in a single time series as well as across other time series. Thus, having a proper aggregation and selection criterion is needed to narrow down the list of potential motifs to those that are most likely to be important or informative and lead to a better interpretation and accuracy.
%about the classifier
 %There are several key aspects when tackling the classification problem. 
 We need to develop a reliable and  interpretable classifier   to distinguish treatment responders and non-responders, with  as low-dimensional as possible representation of the input space. 
 Another desired property of the classifier should be its scalability, meaning it can handle large, high-dimensional datasets and can be easily applied to new patients. This requires taking the computational efficiency of the motif discovery algorithms into account, as well as ensuring that the classifier can be easily integrated into existing psychiatric workflows and systems. %In order to achieve the mentioned properties, the classifier should have an appropriate feature space that is derived from the identified motifs, which play a major role in differentiating between the two patient groups.
% what are the research questions
%This objective consists of several subtasks. We have to effectively extract motifs from multivariate EEG time series data from depression patients and identify the most informative and relevant motifs. This means that we need a selection approach for choosing a subset of the discovered motifs that can be useful for this classification problem. In order to achieve high predictive performance and interpretability, an appropriate feature space that is derived from the identified motifs is needed. This includes investigating how different motif-based feature engineering approaches compare in terms of distinguishing between patient groups, i.e. predictive performance.
% contributions?
%\vspace{-0.3cm}
%\subsection{Main contributions of the work}
%\label{1.3}
%We focus on the following objective,
%\textit{how can we extract motifs that exclusively characterize both groups of respondents and non-respondents from patient EEG signals and develop a binary classifier.}  

\begin{figure}[h!tb]
\centering
\small
   \begin{minipage}{0.54\textwidth}
     \centering
     \includegraphics[width=0.44\linewidth]{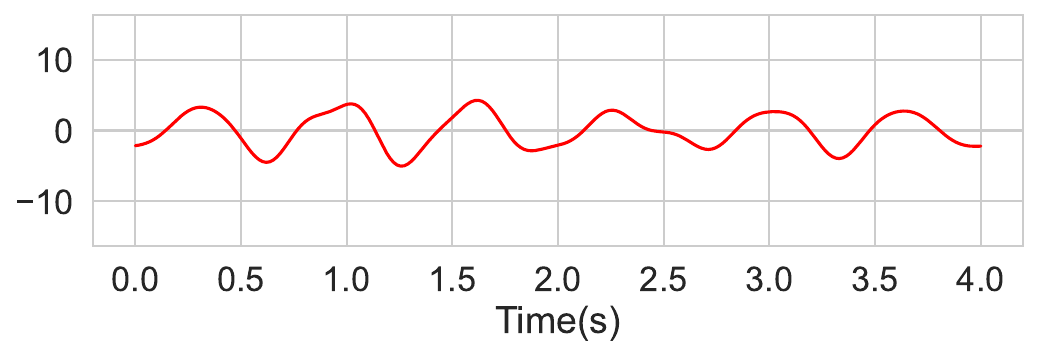}
     \hspace{0.3cm}
     \includegraphics[width=.44\linewidth]{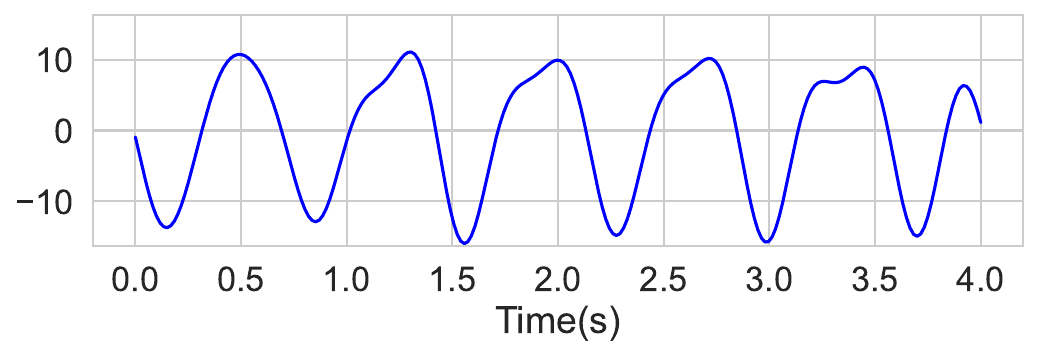}
   \end{minipage}   
   \hspace{1.1cm}
 \textmd{Dataset: MDD} 
   \begin{minipage}{0.54\textwidth}
    \centering
 \includegraphics[width=0.44\linewidth]{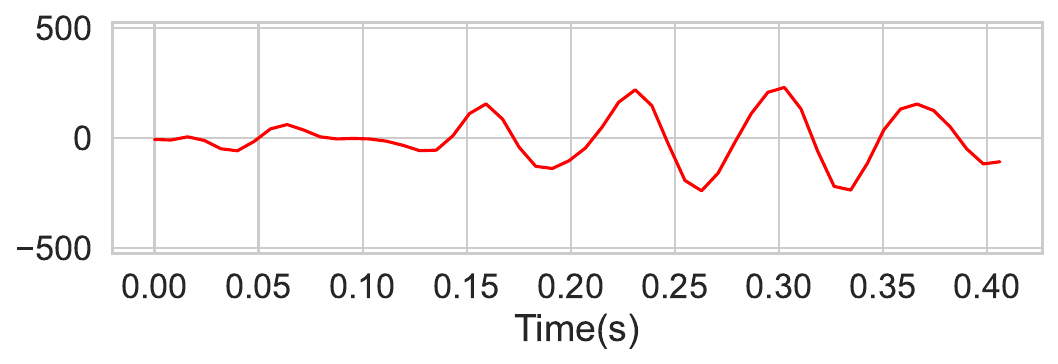}
      \hspace{0.3cm}
     \includegraphics[width=.44\linewidth]{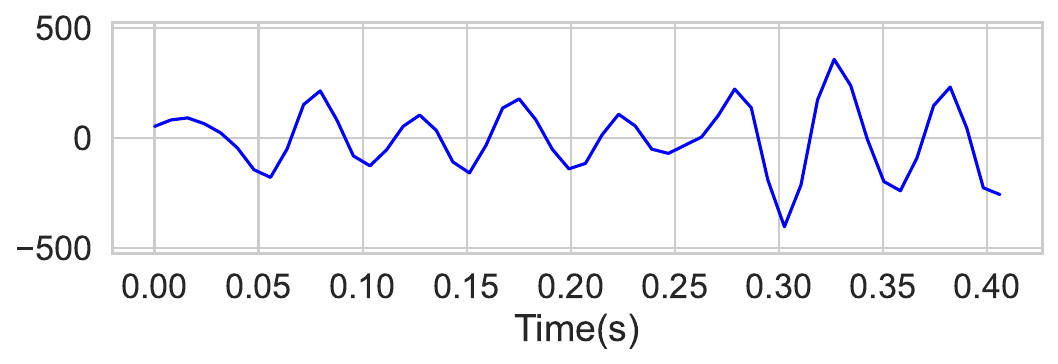}
   \end{minipage} 
 \hspace{0.1cm} \textmd{Dataset: Schizophrenia}
  \begin{minipage}{0.54\textwidth}
     \centering
     \includegraphics[width=0.44\linewidth]{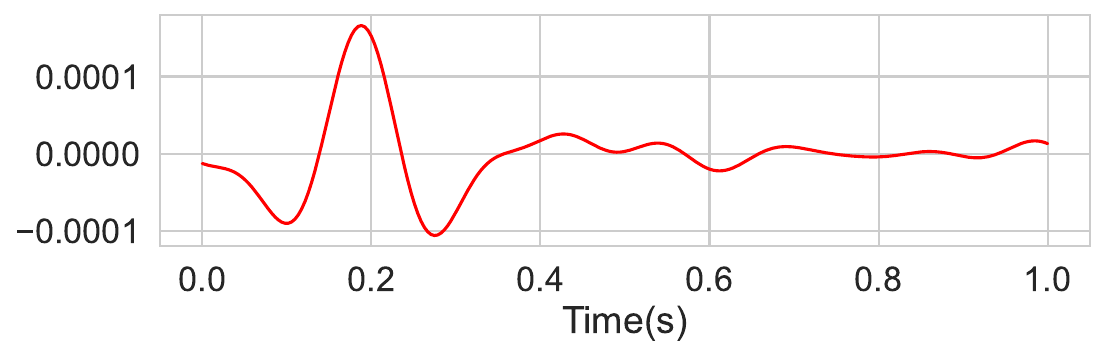}
           \hspace{0.3cm}
     \includegraphics[width=.44\linewidth]{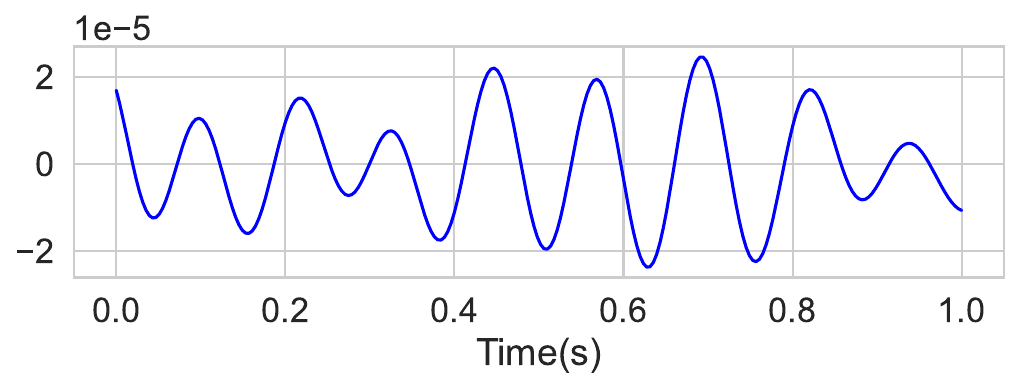}
   \end{minipage} 
      \hspace{0.9cm}
  \textmd{Dataset: Seizures} 
  \begin{minipage}{0.54\textwidth}
     \centering
     \includegraphics[width=0.44\linewidth]{figures/motif_mdd_beta_class0_4.pdf}
        \hspace{0.3cm}
     \includegraphics[width=.44\linewidth]{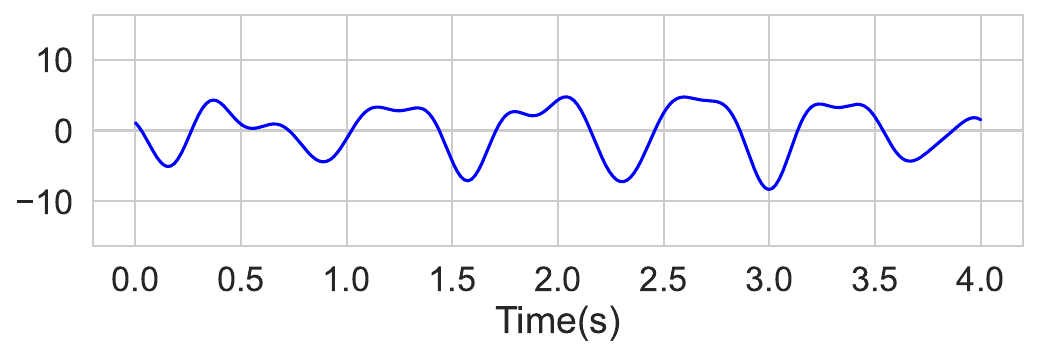}
   \end{minipage} 
         \hspace{0.5cm}
  \textmd{Dataset: Alzheimer's} 
  \begin{minipage}{0.54\textwidth}
     \centering
     \includegraphics[width=0.44\linewidth]{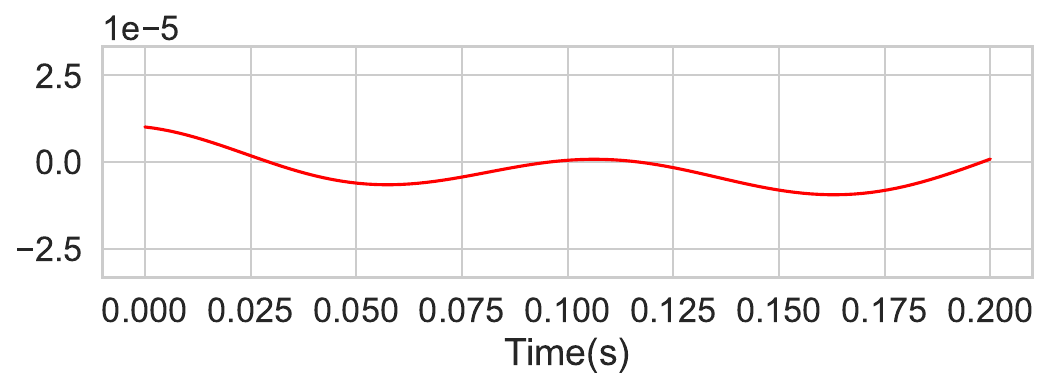}
           \hspace{0.3cm}
     \includegraphics[width=.44\linewidth]{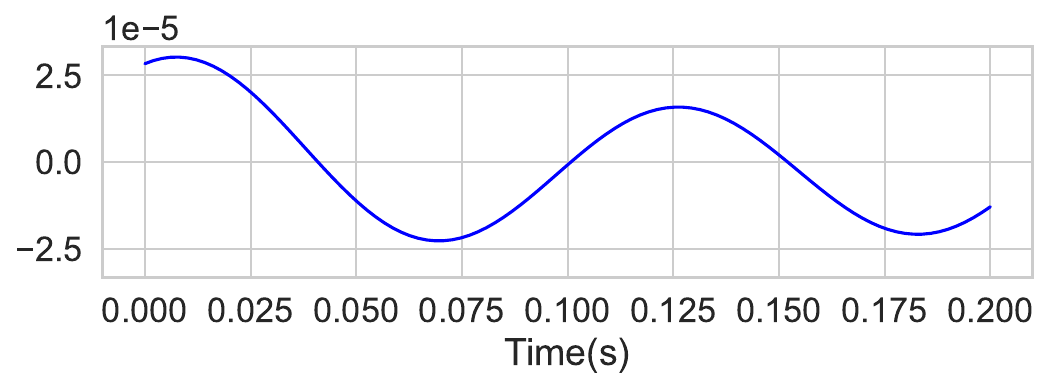}
   \end{minipage} 
         \hspace{0.4cm}
 \textmd{Dataset: Dementia} 
    \caption{Example of motif pairs for each label (label 0, in red  - non-responders  for MDD and  healthy subjects for other sets, and label 1 in blue - responders for MDD and pathology for other sets) from each dataset.}
    \label{motif_pairs}
\end{figure}

\vspace{0.2cm}
Our main contributions can be summarized as follows:

%\vspace{-0.42cm}

\begin{itemize}
    \item We propose a novel framework
including extracting discriminative motifs from EEG data and their usage in classification problem. 

%. The relatively large size of this dataset addresses a common challenge in this area of research, where datasets are typically smaller, providing a valuable opportunity to examine the method's performance on a broader sample of patients.
    \item We analyze the EEG recordings within alpha, beta, and theta frequency bands in separate experiments. 
    %Each experiment workflow consists of the application of one motif discovery algorithm, where a large number of motifs are extracted. 
    \item To obtain the final set of meaningful motifs, we propose a motif selection criteria to rank the motifs based on their higher likelihood of distinguishing between the classes. 
   % \item We use two versions of support vector machine (SVM), decision tree, random forest, logistic regression and multi-layer perceptron as classifiers.
   \item We applied our framework  to classification  of  depression treatment responders  and non-responders, as well as to classification in other
psychiatric EEG datasets, namely with patients with schizophrenia symptoms,  patients with intractable
seizures,  Alzheimer disease and dementia.
Figure~\ref{motif_pairs} illustrates the examples of motifs pairs selected  by our framework  for each  class label  from each dataset.
\item In all data
sets we achieved high classification precision. 
    \item To our best knowledge, our method is the first one using motifs to predict the treatment response of depression patients and in depression diagnostics in general.
\end{itemize}

\section{Related Work} 	
\label{literature_overview}

%In this section, we present a summary of the classification challenge involving EEG signals in MDD, which can be found in Section \ref{2.1}. Subsection \ref{2.1.1} specifically addresses the classification of treatment response in MDD. Lastly, Section \ref{2.2} contains an overview of the current motif discovery methods and their properties.

\subsection{Classification Using EEG Data in MDD}
\label{2.1}

Recently,  researchers have been investigating  EEG signals to discover characteristics or patterns related to a certain psychiatric disease. Such findings are often referred to as biomarkers, which can be useful in identifying the presence of a disease, in discovering pathophysiological mechanisms, and predicting outcomes of treatment. 
These classification problems require a suitable feature extraction technique that can discover informative features from the EEG signals.
%Since the  EEG signals are non-stationary, non-linear, non-Gaussian, and are used for  multiple channels, the process of feature extraction presents a major challenge. 
According to a review \cite{discriminative2021} that investigated EEG features for detecting MDD,  EEG signals have been shown to have effective discriminative power over differentiating MDD patients and healthy subjects, 
%Significant differences in frequency band power have been shown to be useful for diagnosis, however, the findings are inconsistent even within the same frequency band. One consistent result is the hyper-activation of the right prefrontal cortex in MDD, associated with withdrawal behaviors. In regard to time-related changes, MDD patients also exhibit different values in event-related potentials (ERP), such as lower amplitudes or shorter latency. Differences in ERPs related to stimuli type and feedback are also found, although the results are not conclusive. Complexity metrics indicate higher values of nonlinear parameters in MDD, reflecting fractal and unpredictable characteristics of the data. MDD is associated with increased EEG coherence, indicating heightened neurophysiological connectivity, and abnormal graph properties. However, summarizing the results and findings obtained across different papers, the authors conclude that specific brain areas linked to changes in EEG activity remain uncertain and require additional exploration.
In clinical research, EEG signals were initially utilized primarily for visual analysis of their spatial and temporal properties. More recent research focuses on investigating these properties within specific frequency bands, with the computation of more complex features. Čukić et al. \cite{vcukic2020successful} present the effectiveness of Higuchi’s fractal dimension  and Sample Entropy  and explore multiple classification models, reaching an accuracy in the range of 90.24\% to 97.56\% on data set of 23 patients. 
%Besides the prediction of the presence of depression, EEGs have been utilized for other cases of classification.
Bučková et al. \cite{buvckova2020predicting}  used the same dataset of EEGs that we use but  their objective was to classify biological sex from the EEG recordings. 
%The aim is to test the hypothesis of the presence of higher beta power in women compared to men, as well as in the presence of depression. The authors have obtained the best performance with a convolutional neural network model, achieving 81\% accuracy. 
%Deep learning approaches have been also used in the diagnosis of depression disease.
A review \cite{cipriani2018comparative} of deep learning approaches for diagnosing MDD and bipolar disorder using EEG signals emphasizes that EEG-based methods have substantial potential for assessing and monitoring both diseases.
%However, they still remain "black boxes", as their lack of interpretability poses a significant challenge for physicians, who may be unable to explain to patients how their diagnosis was determined. %This prevents the community from further developing reproducible and deterministic protocols and achieving clinically useful results.

\vspace{-0.1cm}

\subsection{Predicting Response to MDD Treatment}
\label{2.1.1}

Working with EEG signals often implies dealing with high- dimensional data, due to the number of electrodes, duration of recordings, and sampling rate. 
%A study \cite{highdim2021implications} concludes that as the dimensionality of a problem increases, the number of reproduction events decreases exponentially, leading the system to enter an infinite cover-delete cycle. As a result, it is more difficult for the classifiers to produce highly general and accurate rules beyond a certain number of dimensions. 
Althian et al. \cite{highdim2021impact} 
investigated the impact of dataset size on the performance of common
 machine learning classifiers in the medical domain. They observed  that  the overall performance of classifiers depends more on how much the extracted features  represent the data  rather than the data set size.
%Moreover, they argue that a robust model that %utilizes a limited dataset does not guarantee the %best performance compared to other models.
  \cite{christinaeeg} and \cite{christinaeeg2}  computed Granger-causal networks in different frequency bands on the same MDD patient data set we use. %The training was done for each gender separately. 
  The classifier with the highest evaluation score was the decision tree classifier, with an F1 score  0.61.
%The current approaches to address the issue of treatment response in MDD using EEG signals mainly focus on developing and utilizing suitable feature extraction methods. They predominantly focus on the static property of the signal, e.g., statistic properties, application of non-linear methods, time and frequency domain features, etc. 
\cite{alexeeg}
 used  entropy-based measures on the same MDD set with classification accuracy  0.7 and an F1 0.65.
 Mueen et al. \cite{dame2009finding}  proposed to use  motifs as features for classification in disk-resident data.
 %The authors set up a disk-aware algorithm to find fixed-length motifs in multi-gigabyte databases containing massive amounts of time series which detects motifs of a specific length in the extracted independent components of the signal. 
 %They extract the features using a similarity measure between motifs and use the computed distance between motifs as features. 
 To our best knowledge,  motifs as building blocks of features from EEGs  have neither been used for predicting the diagnosis of depression, nor in predicting the outcome of the anti-depressive treatment.
%In the scope of the  same research project, "Learning Synchronization Patterns in Multivariate Neural Signals
%for Prediction of Response to Antidepressants" an don the same data set
%\cite{learningeegproject, czechinstituteoftechnology}, 

%The result might be influenced by  that separating the dataset by gender meant using less data for both trainings, hence the classifier could not learn to generalize well. 

\subsection{Classification Using EEG Data for Other Psychiatric Diseases}
There is a large number of publication on  epileptic seizure detection by machine learning classifiers, see e.g. the recent review \cite{siddiqui}. \cite{sairamya} used wavelet transform and deep network to   classify in 16-channel EEG schizophrenia data basis on 84 patients from \cite{schizophrenia}  (45 with schizophrenia and 39 healthy).  \cite{miltiadous} classifies  23 patients with fronto-temporal dementia, and 29 healthy patients and uses six common classifiers with the accuracy of  78.5-86.3\% of correct detection. To our best knowledge, we are not aware of any publication using motif discovery on the data bases with these diseases.
We used these data sets and tested our method on them. The results can be found in Section~\ref{othersets}.

\subsection{Motif Discovery Algorithms}
\label{2.2}

%% Note: Rewrite eveything, start right away with the most recent methods (include competitors).

% overview of motif discovery algorithms, what are their properties.

%Motifs are a recurring theme in the literature on time series analysis and have been variously referred to as patterns, trends, sequences, shapes, episodes, or frequent subsequences, among other names \cite{torkamani2017survey}. %Survey on time series motif discovery
Motif discovery algorithms vary based on the specific application. Some of the algorithms can identify exact or approximate motifs,   motifs with fixed lengths, or variable lengths; Some algorithms find  motifs in univariate and some in multivariate time series.
%Dealing with either univariate or multivariate time series data is also important to consider when finding motifs, as multivariate data is more complex to analyze and identify meaningful patterns across different channels. 
%In short, when choosing a motif discovery algorithm, it is crucial to consider the algorithm's properties and ensure they align with the specific requirements of the application at hand.
Here we make a short overview of recent motif discovery algorithms and their applicability to our classification problem. 
Exact motif discovery refers to the process of identifying recurring patterns or motifs in time series data, where the discovered motifs are identical or nearly identical matches. %The state-of-the-art algorithms for motifs  discovery of a fixed length are the MK \cite{mk2009exact} and the QuickMotif \cite{quickmotif2015}. Their shortcomings   are that in less ideal situations, both algorithms can degenerate to brute-force search and require more memory than more recent methods.%
Yeh et al. introduced an algorithm called STAMP \cite{stamp2016matrix} which utilizes a fast similarity search algorithm to find exact fixed-length motifs in a time series of length $l$.
%, with time complexity $O(l^2log(l))$.
The authors further introduce STOMP \cite{stomp2016matrix} with  reduced complexity.
%$O(log(l))$. % Both algorithms have time and space complexities that are independent of the given motif length. However, despite its slower computational time, STAMP is often preferred over STOMP due to its rapid convergence. 
%In most cases, running STAMP to a partial completion is sufficient to obtain a precise estimation of the desired solution. 
The algorithm SCRIMP++ \cite{scrimp2018matrix} combines the features of both STOMP and STAMP.
Schäfer et al. in \cite{schafer2022motiflets} proposes an algorithm to find motifs, referred to as \textit{Motiflets} - the set of exactly $k$ occurrences of a motif of length $l$ with minimal maximum pairwise distance.  The authors argue that setting $k$ is more intuitive and easier  than setting  the motif length $l$ or the distance threshold between the motif occurrences $r$. The paper proposed two extensions to learn the input parameters $k$ and $l$ from the data and applied  it in the experiment with EEG sleep signals. Two largest motif sets were found, which correspond to the well-known motifs in sleep EEG data (K-Complex and sleep spindles). For the superior precision  over  the compared more recent  methods for this task, namely VALMOD \cite{valmod}, EMMA \cite{emma}, SetFinder \cite{setfinder} and Learning Motifs \cite{learningmotifs}, and  the convenience of automatically determining suitable values for the input parameters, we use this method in our motif discovery step. %of the workflow.

\section{Background}
\label{background}

%FROM HERE
%In this section, we provide a more detailed description of the motif discovery algorithms that are used in this thesis. In Section \ref{3.1}, we present the theory behind the matrix profile computation and how it is used within the SCRIMP++ algorithm, while in Section \ref{3.2} we extend this theory to the discovery of consensus motif by the Ostinato algorithm. 

\subsection{Motiflets}
\label{3.1}

% general, maybe some definitions
Discovering k-Motiflets can be achieved either by using the exact or approximate solution. Both solutions take as input the motif length $l$ and the number of its occurrence $k$ (previously determined by the proposed extensions of the method), and compute the k-Motiflets. Both the exact and approximate algorithm  first generate the candidate motif sets by pairing each subsequence of a time series $T$ with its non-overlapping $k$-1 nearest neighbors (NNs). Then the extent $d$ is calculated for each of these sets, representing the maximum pairwise distance among the subsequences. 
The exact solution considers all subsets of subsequences of $T$ of size $k$ together with a heuristic on pruning to reduce the number of candidate sets, however, it obtains an exponential complexity on $k$. The approximate solution represents a greedy approach which computes the z-normalized pairwise distance matrix and applies pruning to limit candidate sets based on an upper bound on the current best extent $d$. This algorithm assumes that the NNs of a core subsequence are part of the final k-Motiflet, and returns the subsequences with the smallest $d$. For every candidate set where at least $k$ subsequences are within $d$, it extracts the closest non-trivial subsequences and calculates their pairwise extent $dist$. Then, $d$ is updated in case it is larger than the newly obtained $dist$. 
In the worst case, the complexity of the approximate solution is $O(k n^2) + O(n k^2)$ where $n$ is the length of time series. The best case regarding complexity would be the case where the first subsequence is the top k-Motiflet and would allow pruning all further computations, leading to a complexity of $O(n^2) + O(k^2)$ \cite{schafer2022motiflets}.

\noindent Determining meaningful values for $l$ and $k$ are based on the extent function (EF), where $EF(k)$ returns the extent of the top k-Motiflet set, $S_k$, with  fixed length $l$. 

% note: add example photo

\subsection{Learning meaningful \textit{k}}
\begin{figure}[h!]
%\vspace{-0.2cm}
\includegraphics[width=10cm]{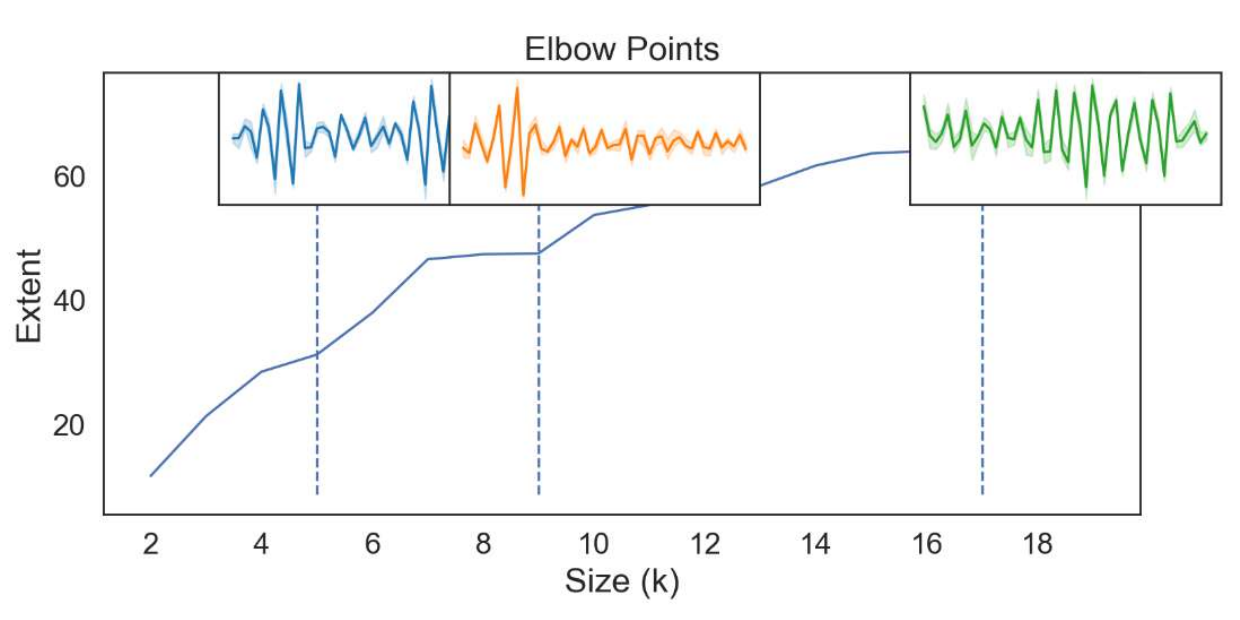}
\centering
\caption{Example plot of the elbow method of choosing the top motifs from the electrode's alpha band with a motif length of 2s, using the \textit{plot\_elbow} \cite{motifletsrepo}. } \label{elbow_point_k}
\end{figure}

Looking at the outputs of EF over increasing values of $k$, allows for inspecting the points with a significant increase of the slope of the line between consecutive values of $k$, known as \textit{elbow points}. These points indicate that considering the next larger value of $k$ causes a significant increase in the extent. The potential values for $k$ are determined by using a threshold for detecting the elbow points. An example of the elbow method is shown in Figure \ref{elbow_point_k}, where the elbow points in 5, 9 and 17 reveal the top motifs.

%\vspace{-0.2cm}

\subsection{Learning meaningful \textit{l}}
To determine $l$, the authors propose computing the normalized area under EF curve, $AU\_EF$, which considers the list of $k$-1 extents for a given length $l$ and the number of elbow points. The intuition is that elbow points are followed by typically steeper slopes, resulting in a larger $AU\_EF$. Thus we want to choose the $l$  minimizing the $AU\_EF$ score. 

% note: add example photo

%\vspace{-0.1cm}

\section{Proposed Method}
\label{methodology}
Our primary objective was to propose  a method workflow using k-Motiflets for a reliable classification in the most challenging data basis of MDD patients,  compared to  four mentioned  data sets with other  psychiatric conditions.
  We  describe first  this  EEG  data basis and then  the method workflow. The details about the  preprocessing and transformations of the data are in Appendix \ref{preprocessing_eegs}. 
%\subsection{EEG Database}
%\label{data}
The patient database 
%recorded and provided by the Czech Academy of Sciences within the "Synchronization" project,  
contains  recordings of 176 patients who are treated for MDD with antidepressants. We use  10-min sessions of 19-channel EEG recordings per patient  (resting state with the eyes closed), 
%the first one  recorded before the treatment, and the second one  
 recorded on the 7th day of the treatment. The electrodes in the 10-20 standard EEG setup are placed on the patient's scalp   with channels: Fp1, Fp2, F7, F3, Fz, F4, F8, C3, Cz, C4, P3, Pz, P4, T3, T4, T5, T6, O1 and O2.

%\vspace{-0.3cm}

%\newpage

\subsection{Labels} 

To evaluate the patient's response  to the treatment, psychiatric experts used the MADRS score and determined it before  the treatment, and then on the 28th day after the start of the treatment.  MADRS score ranges  from 0 to 60 indicating the level of severity of the depression.  The response to the treatment is determined based on the score difference between the two evaluations, indicating the level of improvement/worsening of the patient's well-being.
%Hence, the class labels indicate whether a patient is a responder or a non-responder.
A patient is  a responder (Label 1) if the MADRS score in 28th day reduced by 50\% compared to the initial MADRS score, otherwise is a non-responder (Label 0).
%\vspace{-0.4cm}
 The  dataset  consists of 84 responders and 92 non-responders,   48 male and 128 female patients.   

\subsection{Frequency bands} 
%As pointed out in Section \ref{literature_overview},
Some research papers indicate that the alpha \cite{analysis2018} \cite{alpha2021frontal}, beta \cite{buvckova2020predicting} and theta band \cite{analysis2018} \cite{theta2015frontal} contain representative biomarkers for depression detection or depression treatment outcome, hence the experiments are conducted on these frequency bands of each of the electrodes.  We used the Python MNE library \cite{mnegramfort2013meg} to extract the frequency bands theta (4-8 Hz), alpha (8-12 Hz), and beta (12-30 Hz). A short segment of the extracted bands of an electrode can be found in Figure~\ref{frequency_bands} in the Appendix \ref{preprocessing_eegs}.
The complete workflow from motif extraction to building a classifier is conducted for each frequency band separately and consists of several key steps: motif discovery for motif extraction using k-Motiflets, then selecting the motifs with the highest discriminatory power using the motif selection criteria, and then preparing the feature matrix for the classification step.

\subsection{Motif Discovery}
\label{motif_discovery}

The main and most extensive part of the whole workflow process is the motif discovery for different frequency bands: alpha, beta and theta. We use the k-Motiflets algorithm together with its proposed heuristics for determining the motif length $l$ and occurrence $k$. 
In the first step of the motif discovery, we conduct the search for the most suitable motif length $l$. This is done by using the \textit{find\_au\_ef\_motif\_length} \cite{motifletsrepo} method, which takes the list of potential motif lengths $L$ as input, together with the maximum value of $k$, $k_{max}$. In order to pick an appropriate list of motif lengths for determining $l$, we carried out several initial experiments with different lengths, and settled for list of potential motif lengths ranging from 0.2 seconds to 8 seconds. As for $k_{max}$, we used 20, in accordance to the experiments conducted by the authors. Once $l$ is obtained, we run the \textit{search\_k\_motiflets\_elbow} \cite{motifletsrepo} method, which for each $k$ up to $k_{max}$ computes the EF and returns the list of motif set candidates for each $k$, together the distances for each occurrence, as well as the elbow points which point to the best candidates.
The simplified pseudocode of this approach is presented in Algorithm ~\ref{motiflets_extract}.

\RestyleAlgo{ruled}
{\small 
\begin{algorithm}[H]%[ht]
\vspace{0.2cm}
\caption{\textbf{Motif extraction with k-Motiflets}}\label{motiflets_extract}
\KwData{$E^1, ... ,E^{19}$, where $E^i$ is a set of all $i^{th}$ electrodes of all patients, $L$ - set of potential motif lengths, $k_{max}$ - maximum number of motif occurrences}
\KwResult{motifs[] - a list of extracted motifs, ks[] - a list of their corresponding $k$}
$motifs \gets []$\; \\
$ks \gets []$\; \\
\For{$i=1$ to 19}{
    \For{each $e$ in $E^i$}{
        $l \gets find\_au\_ef\_motif\_length(e, L, k_{max})$\; \\
        $dists, candidates, elbow\_points \\ \gets search\_k\_motiflets\_elbow(e, l, k_{max})$\; \\
        \For{each $motiflet$ in $candidates[elbow\_points]$}{
        $k \gets len(motiflet)$\; \\
        $idx \gets motiflet[0]$\; \\
        $motif \gets  e[motif_{idx} : motif_{idx} + l]$\; \\
        $motifs.append(motif)$\; \\
        $ks.append(k)$\;
        }
    }
}
%\hline
\end{algorithm}
}

% note: example of the elbow plot.

\subsection{Motif selection criteria} 
\begin{figure}[H] %[!ht]
   \begin{minipage}[!b]{\linewidth}
     \centering
     \includegraphics[width=.4\linewidth]{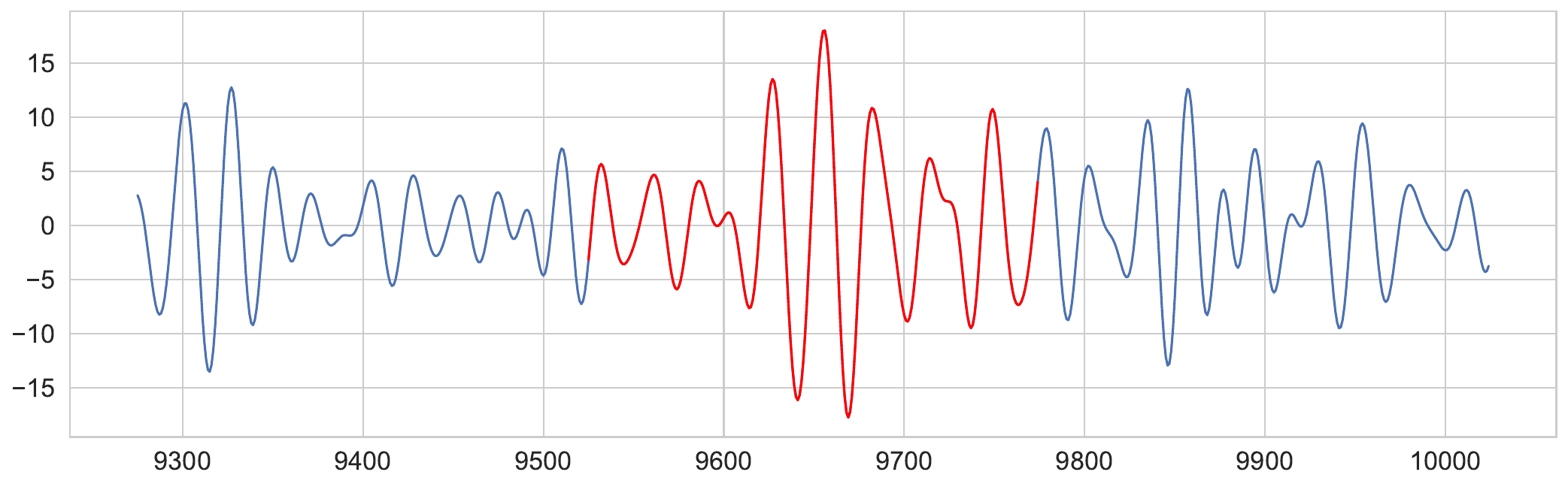}
   \end{minipage} \\
   \begin{minipage}[!b]{\linewidth}
     \centering
     \includegraphics[width=\linewidth]{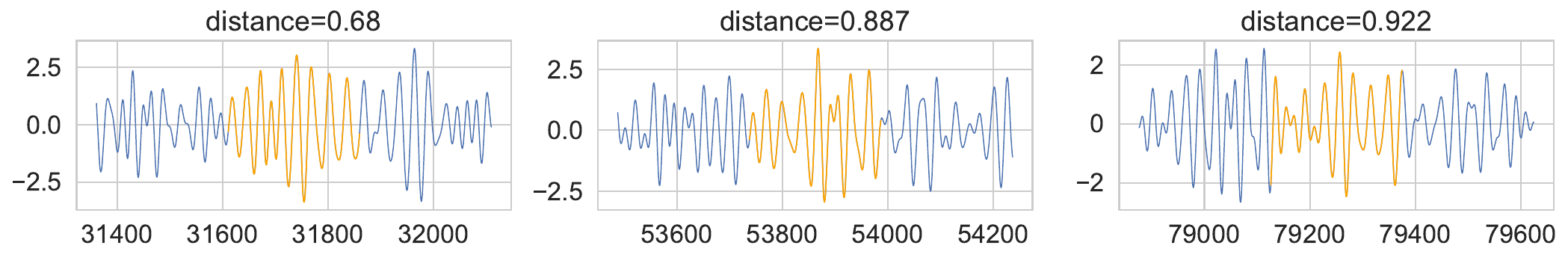}
   \end{minipage} 
    \caption{An example output of \textit{match} returning the distances to the best matches (orange) to a motif (red).}\label{motifsmatch_ex1}
\end{figure}

After the motif discovery step, we obtain a large set of motifs.  To  filter out motifs that do not have any discriminatory power to differentiate between responders and non-responders, we need to check which motifs appear to be typical for one class, i.e., are not common patterns for both classes.
To assess this, we need to   check the presence of a motif within a given signal. This involves computing the closest subsequence within the signal, as well as identifying the nearest neighbors to this subsequence - we referred to these closest subsequences to the motif as the \textit{motif matches}. For identifying the closest match of a motif within a time series, we use the  $match$ method within Python library STUMPY \cite{law2019stumpy}. The method uses the MASS \cite{stamp2016matrix} algorithm and provides a fast way to compute the sliding window dot product between the query subsequence (in our case the motif) and all the subsequences in the given time series. We use the method to obtain a list of $k$ closest matches, including the z-normalized Euclidean distance between the motif and the match, see an example in 
Figure~\ref{motifsmatch_ex1}. The maximum distance for which a subsequence in time series $T$ of length $n$ is considered a match for a given motif $Q$ of length $m$ is defined as: 
$
f(D) = \max\left(\left(\text{{mean}}(D) - 2 \cdot \text{{std}}(D)\right), \text{{min}}(D)\right)
$
where \(D\) is an array with a size of \(n - m + 1\) and represents the distance profile of \(Q\) with \(T\). Hence,  function \(f(D)\) returns at least the closest match. %An example of an output of the method \textit{match}
%(the obtained matches together with their distances) 
%to the motif is  in Figure \ref{motifsmatch_ex1}.

\vspace{0.3cm}

\RestyleAlgo{ruled}

{\small 
%\begin{algorithm*}[th]
\begin{algorithm}%[h]
\vspace{0.2cm}
\caption{\textbf{Compute difference score of a motif}}\label{motif_match}
\KwData{$Q$ - motif, $E^j$ - list of all $j^{th}$ electrodes, $L$ - binary list of corresponding labels, $percentage$ - the percentage of lowest distances to consider}
%\caption{Compute the difference score of a motif}
\KwResult{$score_{diff}$}
$distances \gets $dict()\; \\
$distances[0] \gets $[]\; \\
$distances[1] \gets $[]\; \\
\For{$i=1$ to $len(E^j)$}{
    $electrode \gets E^j[i]$\; \\
    $label \gets L[i]$\; \\
    $distances_{matches} \gets match(Q, electrode)$\;  \\
    $distances[label].append(mean(distances_{matches}))$\;
}
$best\_distances\_0 \gets get\_best(distances[0], percentage)$\;
$best\_distances\_1 \gets get\_best(distances[1], percentage)$\;
$score_{diff} \gets abs(best\_distances\_0 - best\_distances\_1)$\; 
%\hline
\end{algorithm}
}
%\vspace{-0.3cm}

\noindent For checking the motif presence, we have to consider the distances of the subsequences within the motif match to the given motif and decide on a suitable threshold for determining which distance (or average distance) is acceptable for considering the found match represents an instance of the motif.  Finding such a  threshold is difficult, as it may also vary depending on the motif length and the frequency band. Therefore, instead of a threshold, we use the average distances within the match to represent how distant is the found match to the given motif. Motifs with a greater discriminatory power between the two classes are expected to have close matches in one class and more distant matches within the other class. 
To quantify this discriminatory power, we compute the \textit{difference score} in Algorithm \ref{motif_match} by considering the distances of the motif to its matches within each class. If there is a greater difference in the average distances between the classes, that could indicate that the motif is mainly present in one of the classes. 
%\Mag{Considering that a motif could be typical for a subset of patients within one class, we take only a \textit{percentage} of the lowest average match distances from both classes and then compute the absolute difference.} 
For computing the average distance of a motif to its class, instead of considering the average match distances to all patients of the class, we consider a \textit{percentage} of the lowest average match distances. This allows for a low difference score for motifs that are typical for a subset of patients within one class.  
It is important to mention that when computing this score, we consider only the electrodes that correspond to the motif's electrode of origin, i.e. if the motif originates from a patient's $j^{th}$ electrode, we consider all $j^{th}$ electrodes.

Using these scores significantly reduces the pool of motif candidates. In order to have a balanced representation of each class and gender, we choose the same number of motifs for each class-gender combination. Hence, we preserve the best $n$ motifs for each class-gender combination based on the difference score. 

\subsection{Features} 

The feature matrix consists of rows -  patients and the columns - the motifs. Patient $i$ is represented as a $d$-dim vector $v$, where $d$ is the number of motifs and $v_j$ is the average distance of the closest matches to the $j^{th}$ motif. Similarly as above, for computing this value we use method $match$, considering the patient's electrode that corresponds to the motif. Essentially, the feature matrix represents a distance matrix. 

\subsection{Classification}
%Models, feature selection hyperparameters, validation

For predicting the class label, i.e., the treatment outcome of the patients, we use several common classification methods. To combat overfitting, we have chosen simpler methods, whose hyperparameters allow for regularization. Interpretability is also a crucial property in decision-making in the medical domain. Considering both preferences, we selected the support vector machine (SVM) with two different kernels, decision tree, random forest, logistic regression, and multi-layer perceptron from the scikit-learn library \cite{pedregosa2011scikit}. The selection of the hyperparameters of these classifiers can be found in Appendix \ref{hyperparameters}. As for the performance evaluation we use the F1 measure as it accounts for class imbalances  as well as accuracy.
% hyperparameter tunning
%Note: In the case of SVM, we consider two implementations provided by the scikit-learn library: SVC (Support Vector Classification) and LinearSVC (Linear Support Vector Classification), which have different loss functions set by default, and different handling of intercept regularization, as the LinearSVC penalizes the size of the bias. Hence, in the case of SVC, we tune the 'C' and 'kernel' parameters, as it uses the 'l2' penalty by default. In the case of LinearSVC, we explore different penalty settings: 'l1' and 'l2', as the L1 norm imposes a stricter regularization.
% feature selection
To further reduce the feature space and retain the most important motifs, we use the wrapper method Recursive Feature Elimination (RFE) \cite{rfe2002gene}, which iteratively removes the least important feature until the optimal number of features is attained based on a prediction metric. 

%Initially, the estimator undergoes training with the complete set of features, where also the significance of each feature is determined, typically using a particular attribute or callable method. Subsequently, the least important features are removed from the current set. This process is repeated recursively on the trimmed feature set until the desired number of selected features is attained. 
%The number of optimal features to keep is obtained based on the highest prediction score on the validation set. 

%\raggedbottom 

\section{Experiments}
\label{experiments}

 The experiments consist of the identification and selection of discriminatory motifs for a concrete data set and training and evaluation of the classifiers.
 Our Python code can be found under \url{https://anonymous.4open.science/r/motif-discovery-eeg-EF02/}.
 %to distinguish between the responders and non-responders. 

\subsection{Motif Discovery}
\label{5.1}

\begin{figure}[!ht]
\centering
    \textmd{Non-responder - \textit{difference score}: 0.531}\\
   \begin{minipage}[!b]{0.4\linewidth}

     \centering
     \includegraphics[width=.82 \linewidth]{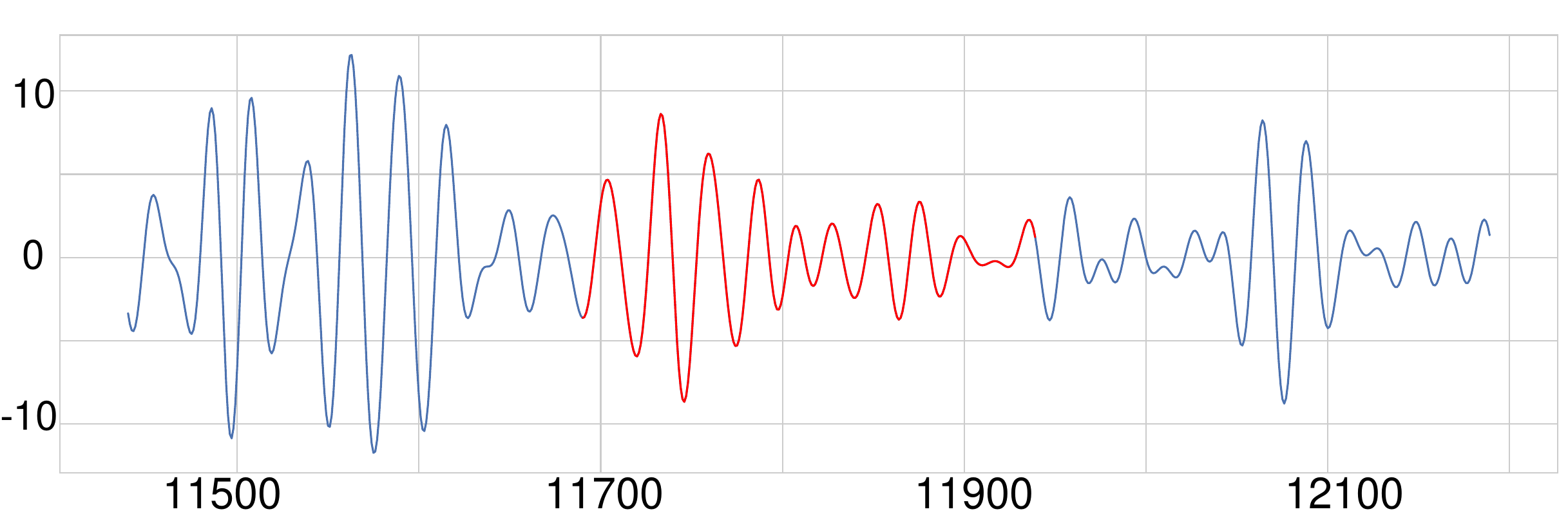}
   \end{minipage} \\
   \begin{minipage}[!b]{0.82\linewidth}
     \centering
     \includegraphics[width=.82\linewidth]{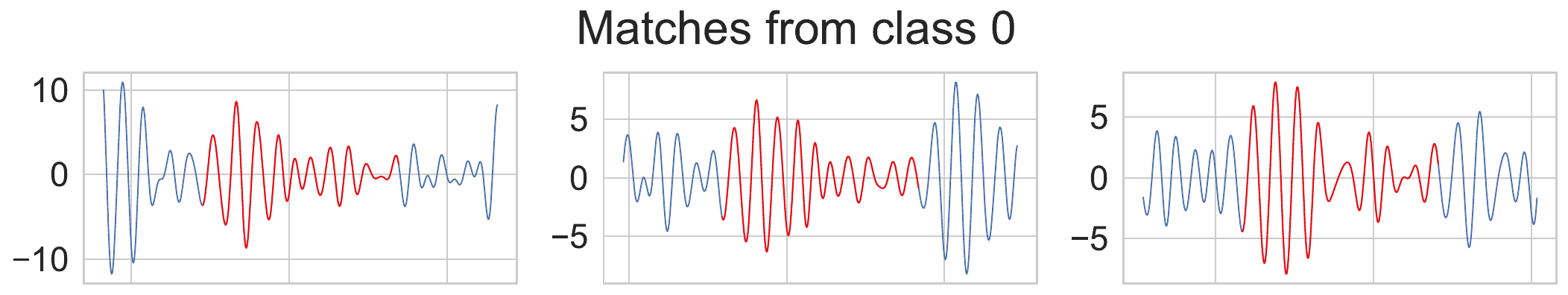}
   \end{minipage} \\
      \begin{minipage}[!b]{0.82\linewidth}
     \centering
     \includegraphics[width=.82\linewidth]{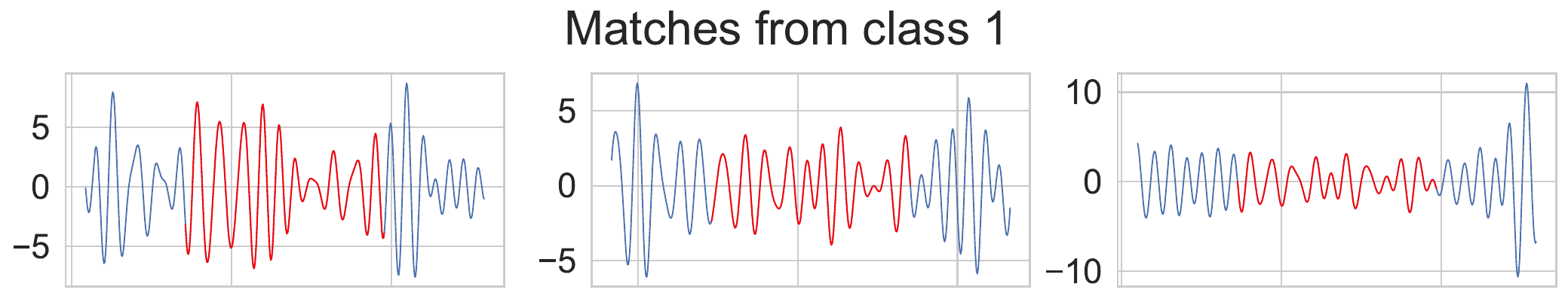}
   \end{minipage}
       \caption{Example of a motif from the beta band and with a higher difference score, together with three examples matches across the two classes. }\label{motifscore_high_1}
\end{figure}

\begin{figure}[H]%[!ht]
\centering
    \textmd{Non-responder - \textit{difference score}: 0.018}\\
\begin{minipage}[!b]{0.4\linewidth}
     \centering
     \includegraphics[width=.82\linewidth]{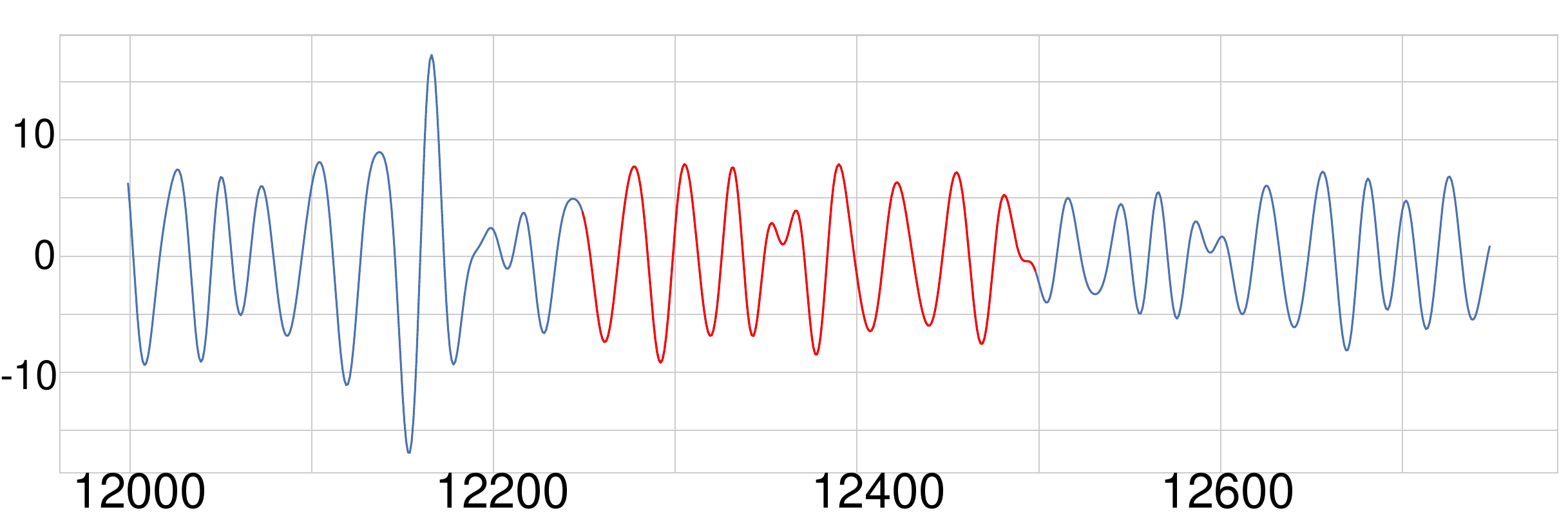}
   \end{minipage} \\
   \begin{minipage}[!b]{0.82\linewidth}
     \centering
     \includegraphics[width=.82\linewidth]{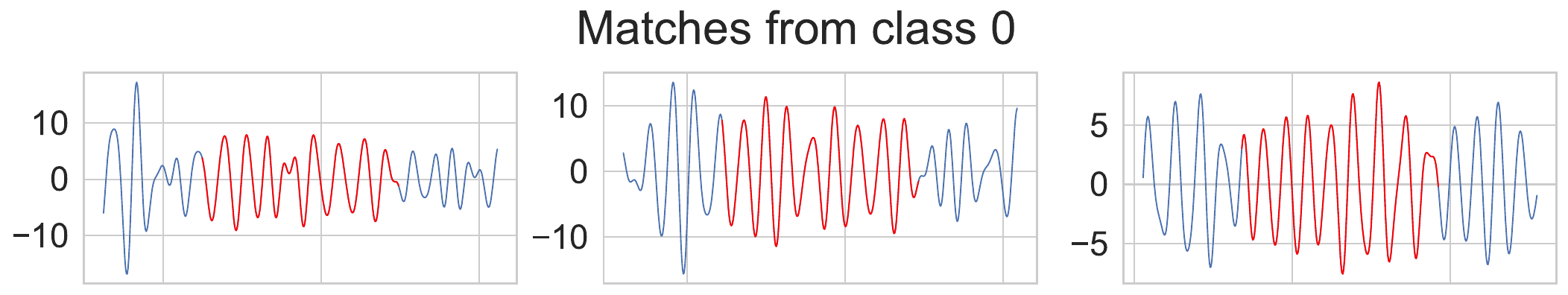}
   \end{minipage} \\
      \begin{minipage}[!b]{0.82\linewidth}
     \centering
     \includegraphics[width=.82\linewidth]{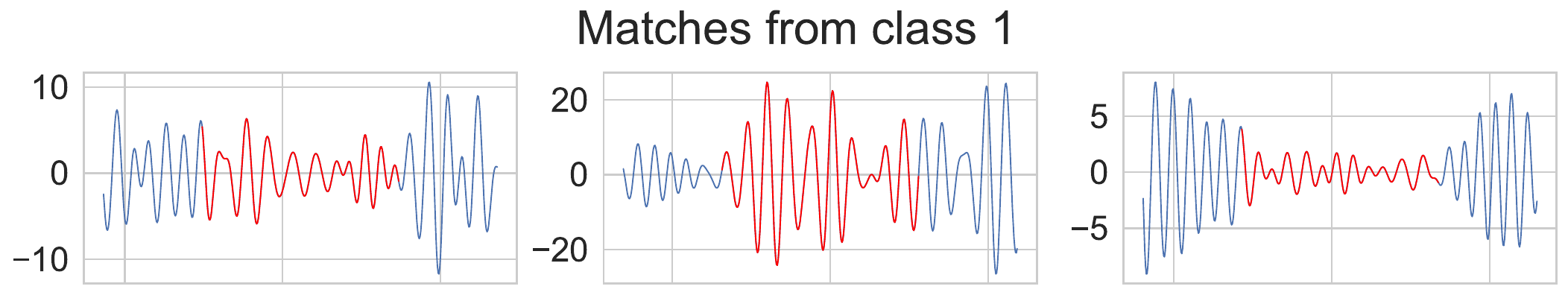}
   \end{minipage}

    \caption{As in  Fig \ref{motifscore_high_1}  with a lower difference score.}\label{motifscore_low_2}
\end{figure}

The experiments were conducted on frequency bands beta, alpha, and theta and  on different motif lengths. Due to the  large number of extracted  motifs for each combination of frequency band and electrode, for each of the motifs we compute the difference score (Algorithm \ref{motif_match}). 
%The following plots refer to the results obtained from the SCRIMP++ algorithm, as both algorithms yield similar results regarding the distribution of the difference score, hence analyzing the results from SCRIMP++ serves the purpose of obtaining a better understanding of the score. 
An example of the score of a motif from class 0 in the depression data basis is depicted in Figure \ref{motifscore_high_1} and Figure \ref{motifscore_low_2}, where a motif for the beta band is shown together with its three closest matches from each class (for readability reasons, we do not show all of the matches within one signal, but rather only the closest match). The motifs with higher difference scores in one class indicate that they have close matches (measured with the normalized Euclidean distance) within the class and more distant matches in the other class. Figure \ref{motifscore_low_2}  is   an example of a motif with a score close to 0, which indicates that there are close matches in both classes, therefore the motif might not be a good candidate for discriminating between the classes.

%\vspace{-0.8cm}

%\vspace{-0.4cm}

\subsection{Classification}
\label{5.2}
We construct a feature matrix with a subset of the obtained motifs, having the largest discriminatory power, i.e. difference score.

\subsection{Imbalance}

The dataset is imbalanced with respect  the class label and gender. Thus, it is of crucial importance to have each combination of class and gender equally represented in the feature matrix. When selecting a smaller subset of the motifs, we choose the same number of motifs for each combination. We would like the number of motifs to be as low as possible, to avoid overfitting, but on the other hand, we would like to include as much information as possible. Hence, after conducting several experiments, we decided to start with the best 20 motifs for each combination (ranked by the difference score) to construct the (initial) feature matrix. We additionally use a feature selection technique to further reduce the feature space and keep the most descriptive motifs.
To obtain a better overview of constructed feature space, we project the obtained patient profiles to a lower dimensional space using Linear Discriminant Analysis (LDA) \cite{lda1936use}. If the projected points from different classes are well-separated in the lower-dimensional space, it indicates that LDA has successfully captured the discriminative information, and the classes are easily distinguishable. This suggests that a simple linear classifier (e.g., logistic regression or linear SVM) trained on the transformed data is likely to perform well. For each of the bands and motif lengths, we visualize the values from the obtained projection to gain further insights into possible linear separability between the two classes. 
%depicting the projected values of an imbalanced feature matrix. 
%\begin{figure}[!htb]
%\begin{minipage}{0.4\textwidth}
 %    \centering
%     \includegraphics[width=.3\linewidth, scaled=0.8]%{figures/boxplot_alpha_imbalanced.pdf}
 %  \end{minipage}\hfill
 %  \begin{minipage}{0.4\textwidth}
  %   \centering
 %    \includegraphics[width=0.7\linewidth]%{figures/stripplot_alpha_500_unbalanced.pdf}
 %  \end{minipage}
 %   \caption{Example of the distribution of the projected values of an %imbalanced training feature matrix of the alpha band, motif length %500.}\label{alpha_imbalanced}
%\end{figure}

\begin{figure}[h!tb]
%\caption*{Band: alpha - Motif length: 1000}
 %  \begin{minipage}{0.4\textwidth}
     \centering
      \includegraphics[width=.35\linewidth]{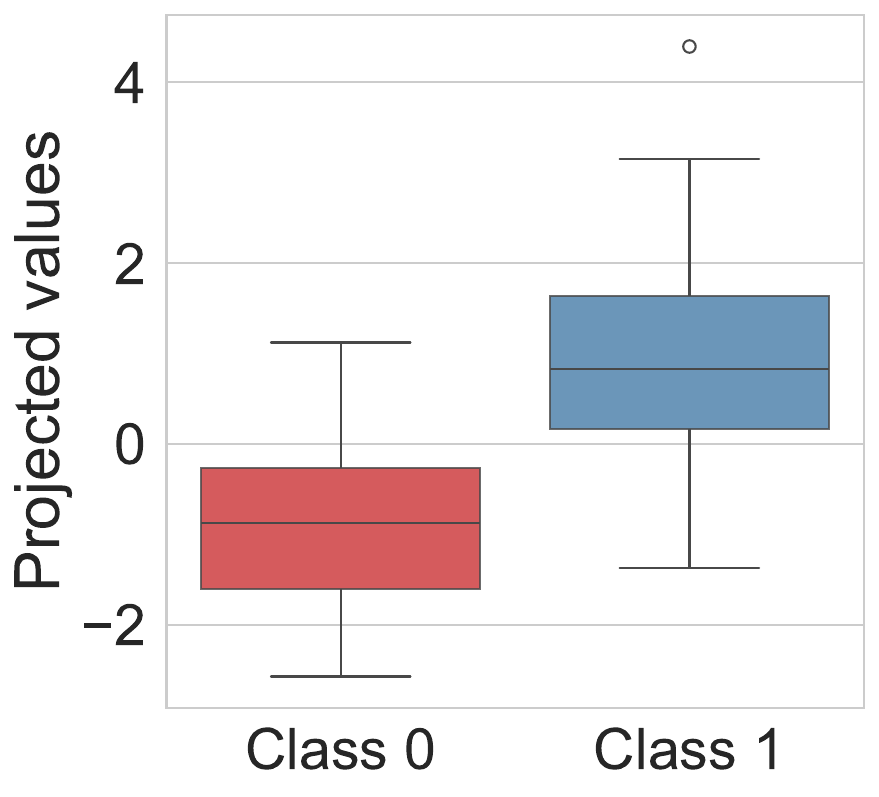}
 %  \end{minipage}\hfill
 %  \begin{minipage}{0.4\textwidth}
     \centering
     \includegraphics[width=0.44\linewidth]{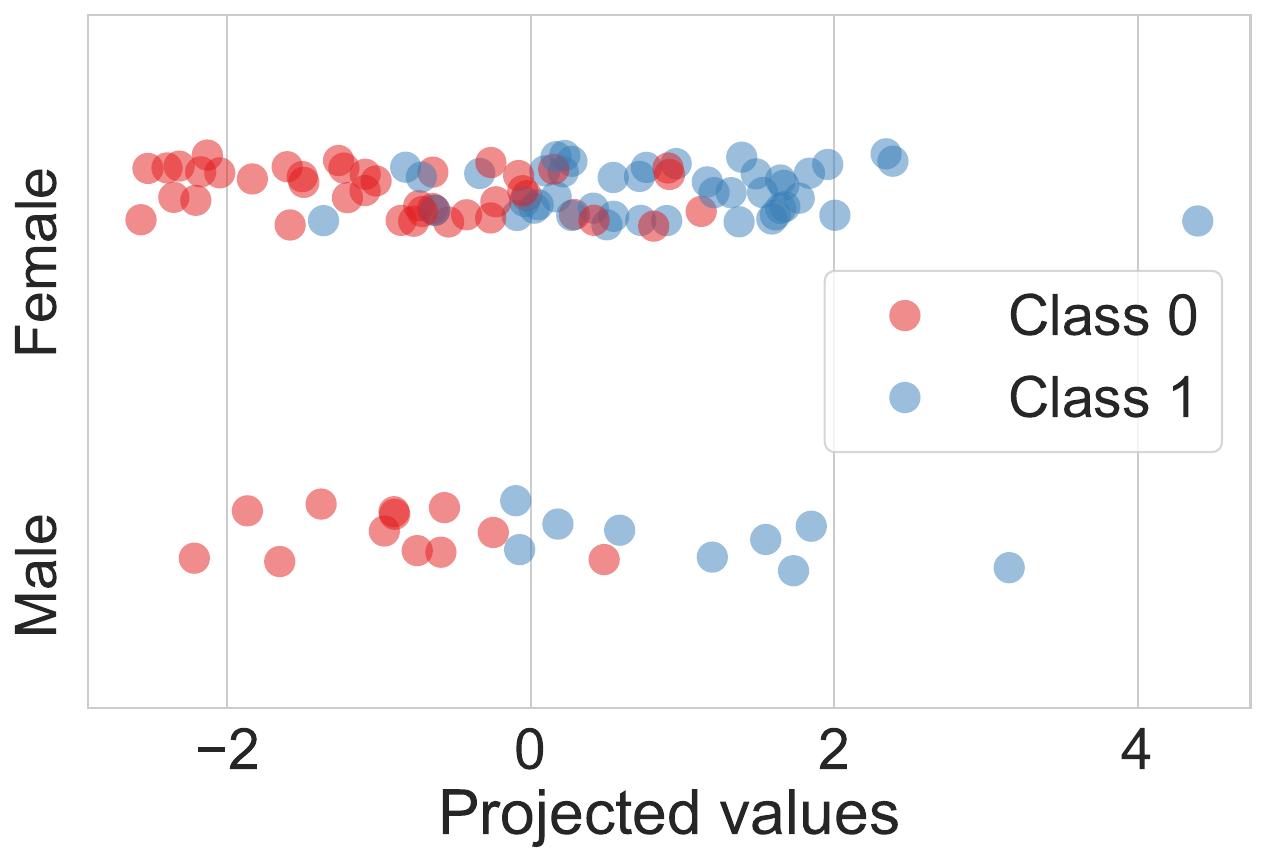}
 %  \end{minipage}\\
    \caption{Example of projected values of the feature matrix for the alpha band.}
    \label{stripplot_all_bands}
\end{figure}

\noindent The boxplot on the left of  Figure \ref{stripplot_all_bands} represents the distribution of the projected values among the two classes, while the scatter plot on the right depicts the projected values, separated by gender. One can see that there is some degree of linear separability between the classes (the corresponding plots for beta and theta bands are in Appendix \ref{feature_matrix_projected_values}).

%\vspace{-0.3cm}

%The plots are consistent with the observation spotted in the previous section, that the shorter motif lengths appear to be suitable for the beta band, while the longer motif lengths could be more suitable for the alpha and theta bands. 

%mention imbalances and the importance of representing each group with the same number of motifs, maybe add a negative example
\vspace{0.2cm}

%% beta

% table of all combinations of the best eval results
% Frequency band - Motif length - Training acc - Training f1 - Evaluation acc - Evaluation f1 
%plots of training accuracies and F1 scores%

\subsection{Evaluation}

\begin{figure}[!htb]
\centering
\textmd{Cross Validation}\\
\begin{minipage}{0.8\textwidth}
     \centering
     \includegraphics[width=0.49\linewidth]{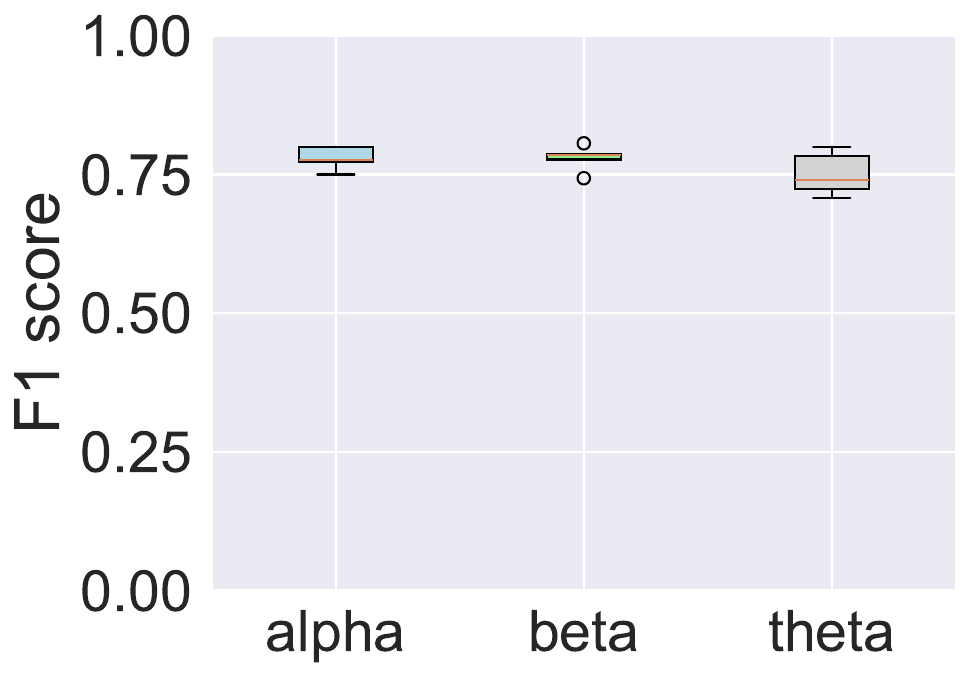}
 %  \end{minipage}\hfill
 % \begin{minipage}{0.47\textwidth}
     \centering
     \includegraphics[width=0.49\linewidth]{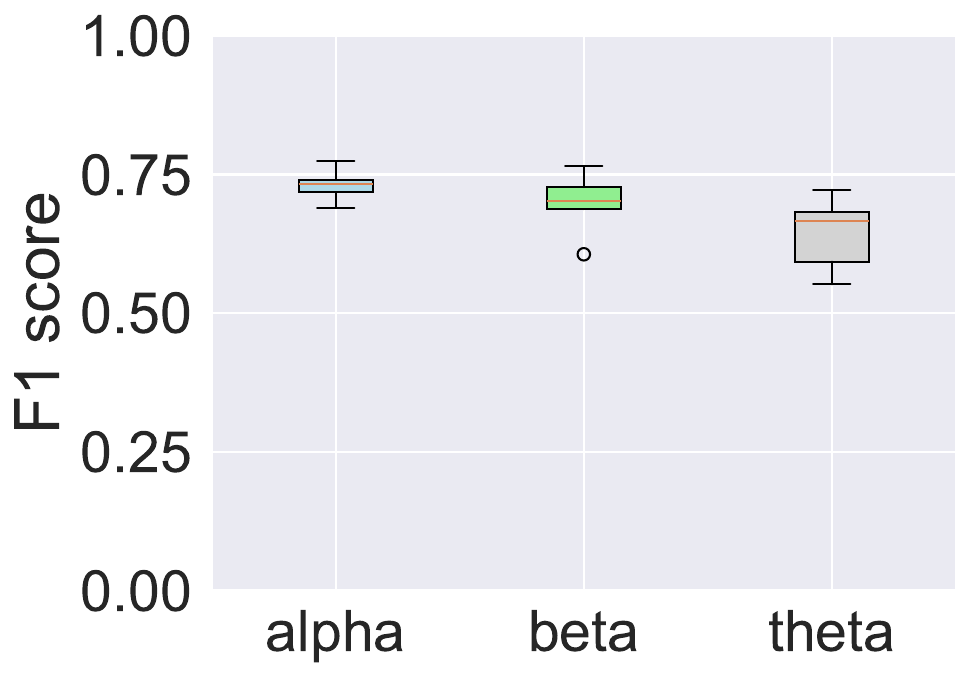}
   \end{minipage} 
\caption{Mean F1  of training (left) and validation (right) for all three freq. bands of the best  classifier.}\label{val_boxplot}
 \end{figure}

%When creating the feature matrices, it is important to stress that the patient profile is constructed from the reduced set of motifs obtained from only the training set (as described in Section \ref{workflow}). The same applies to the patients from  final test set. 
To have a fair evaluation of  the classifiers, we apply a 5-fold cross validation (CV) on the entire dataset of 176 patients and use mean F1 score of the F1 scores obtained from each validation fold.  
%The hyper-parameter tuning is done by 5-fold cross validation on the entire dataset for each classifier. The validation is done by the mean F1 score from the validation folds.
Figure \ref{val_boxplot} shows the mean F1  from CV for each of the three frequency bands, where each box plot depicts the 5 pairs of training and validation sets. One can see that the training F1 scores are quite stable, the classifiers learn and generalize well, and the validation scores behave similarly to the training scores. Balanced scores often result from models that have learned meaningful features and relationships within the data. Similar plots showing the accuracy results for all frequency bands can be found in Appendix \ref{cross_val_accuracy}.

%\raggedbottom

%By examining the variation in the evaluation scores across the K folds, we can have an idea of the classifier's stability and variance in performance.  
%, with some cases with greater variance in the evaluation accuracies. 

Table \ref{train_eval_table} contains the results for each band, obtained by the best classifier (ranked by the highest mean F1 score on the 5-fold CV).  One can see that all three bands have similar scores and also scores are similar for both training and validation sets.

\vspace{0.2cm}

%\begin{center}
\begin{table}[h!]
  \centering
  \begin{tabular}{ccccc}
    \hline
    Band &  \multicolumn{2}{c}{Training} & \multicolumn{2}{c}{Validation} \\
   % \cline{3-6}
     & Accuracy & F1 & Accuracy & F1 \\
    \hline %\hline
    Alpha & 0.785 &	0.78 &	0.731 &	0.722  \\
 %   \cline{2-6}
    Beta & 0.751 & 0.779	& 0.673	& 0.713  \\
 
    Theta & 0.738 &	0.751 &	0.62	& 0.647 \\
    \hline
  \end{tabular}
  \label{train_eval_table}
\caption{Training and validation precision in MDD set for each freq. band - mean accuracy and F1  from CV.}
\end{table}
%\end{center}

\vspace{0.2cm}

\noindent \textbf{Best performing classifier} 
Overall, the best performing model is the decision tree classifier using the decision rule criterion Gini and a maximum depth of 3. The best score is obtained on the alpha band with a mean validation F1  of 0.722 and an validation accuracy  of 0.731.
% \Blu{Should we add evaluation info about the genders? -  a bit difficult given the cross validation} \Mag{so better not}

\vspace{0.2cm}

\noindent \textbf{Feature importance}
 The final feature sets obtained from the feature selection step, contain 8 motifs for the alpha band (4 for responders, 4 for non-responders), and 5 motifs for both beta and theta band (with 1 for non-responders and 4 for responders). The motifs from the electrode  O2 seem to be the most dominant for the alpha band, having 4 of the motifs originating from this channel. The second most important electrode from which originate 2 motifs, is the Fz channel. 

%(for the distribution of counts, see Appendix \ref{importance_channels}, as well as for the beta and theta bands).
%, while in the theta band, we have O2 and Fp2. For the beta band, the electrodes F3, C3, and P3 have  the highest count.

\subsection{Evaluation of the Workflow on Other Psychiatric Data Sets}\label{othersets}
%\Mag{TO DO }

We evaluated the proposed workflow on four publicly available EEG datasets of patients with different psychiatric conditions and their results are summarized in Table \ref{psy_datasets_results}. The dataset of 16-channel EEG of healthy adolescents and those with symptoms of schizophrenia from \cite{schizophrenia} contains 84 patients in total (45 with schizophrenia and 39 healthy).  Our framework found 9 motifs for the healthy and 11 motifs for the schizophrenia group. The dataset from \cite{seizures} contains multiple EEG recordings of 23 pediatric patients with intractable seizures. For classifying the presence of seizures in a recording, we used two recordings per patient (one containing a seizure and one without), having 46 samples in total. Our framework found 1 motif for healthy and 3 motifs for recordings with seizures. The dataset from  \cite{Alzheimer} contains EEG recordings of 88 patients, having 36 diagnosed with Alzheimer's disease, 23 with fronto-temporal dementia, and 29 healthy patients. We used both groups of patients with Alzheimer's and dementia separately to train two binary classifiers that distinguish healthy patients from patients with one of the diseases.
Our framework found 10 motifs for Alzheimer's group and 6 motifs for the control group, and 8 motifs for dementia and 4 motifs for the control group. Results for other frequency bands for each data set can be found in Appendix \ref{other_psychiatric_datasets}. 

\vspace{0.2cm}

%\begin{center}
\begin{table}[h!]
  \centering
  \begin{tabular}{lccccc}
    \hline
    Dataset & Band &  \multicolumn{2}{c}{Training} & \multicolumn{2}{c}{Validation} \\
   % \cline{3-6}
    &  & Accuracy & F1 & Accuracy & F1 \\
    \hline %\hline
    \shortstack{Schizophrenia } & Alpha & 0.927 &	0.931 &	0.820 &	0.821\\
 %   \cline{2-6}
    \shortstack{Seizures} & Theta & 0.834 & 0.831	& 0.787	& 0.784  \\
 
   \shortstack{Alzheimer} &  Theta & 0.868 &	0.847&	0.845	& 0.823 \\
    \shortstack{Dementia} &  Theta & 0.954 &	0.951&	0.87	& 0.865 \\
    \hline
  \end{tabular}
\caption{Evaluation on other psychiatric datasets on the band with the highest mean F1  on CV}\label{psy_datasets_results}
\end{table}
%\end{center}

\noindent One can see from Table~ \ref{psy_datasets_results}  that our method achieves in average comparable or even better classification precision  on all these data sets than on the MDD data set.

\section{Discussion}

The best  classification precision on our  MDD data closely aligns with a recent  meta-analysis review \cite{metanalysis2022predicting} on the prediction of treatment response using EEG in MDD, where the best performing classifier achieves an accuracy score in the range of 0.782 to 0.826. The approaches used in the review paper have a median size of 86.5, and 60\% of them use CV to report on the final results. However a direct comparison of the precision values is problematic, since our database is three times bigger than the data sets in this paper.
%The results reveal that motifs appear to possess some level of discriminatory power over non-responsive and responsive patients, especially within the alpha frequency band. 

\subsection{Comparison of classification precision on MDD to other psychiatric data sets}
We postulate that the relatively lower classification precision on the MDD data set is not in the method but in the data set. Firstly, the MDD data set is compared to  the other data sets larger, which can influence the classification precision. 
%Secondly, the higher classification precision can be due to typical events of the  EEG found as motifs  for these diseases compared to the EEG motifs for the MDD treatment. 
And secondly, the classification precision is also influenced how correctness of the labeling. The class labelling in the MDD data set is done by the MADRS questionnaire, filled in subjectively by the patients on the 28th day of the treatment. On the other hand, the class labelling  in the data sets with schizophrenia, seizures and Alzheimer conditions were done by medical professionals. Thus the lower classification in the MDD set might be related to the fact that a  correct labelling for depression treatment classification is a real challenge.

\vspace{0.1cm}

There are some limitations of the approach:
%LIMITATIONS:
The experimental workflow analyzes motifs up to 8 seconds.
%SCRIMP++ offers significant speed improvements through approximate matrix profiling computation.  
%Although the literature is not consistent in stating which frequency bands are significant for MDD, we selected only three frequency bands, based on recent relevant work.
%, as well as considering the constraints of conducting a large amount of experiments. 
Also, we trained the classifier  using motifs within the same frequency band. The possibility of combining them is limited, since for filtering the motifs that are good candidates we use the proposed difference score. The difference score is based on the distances calculated by the z-normalized Euclidean distance metric, which scales with the length of the motif. In addition, the signals are different within each frequency band, hence in order to be able to do a fair ranking of the motifs across all bands and lengths, the difference score has to be invariant to these two parameters. 

\section{Conclusions and Future Work}
\label{conclusion}

We proposed a novel framework including feature selection, motif discovery and classification   to be used  in general for classification of psychiatric patients based on their EEG.
%prediction of
%MDD treatment outcomes using motifs obtained from the
%patients’ EEGs.
%We tested our framework on classification tasks in five psychiatric conditions, depression treatment, detection of schizophrenia, seizures,  dementia and Alzheimer disease.
The best  classification precision
 was obtained for  data sets  with schizophrenia and dementia which in both cases have training score in F1 and accuracy above 0.92 and by testing above 0.82.
%At the same time, our method achieves in average comparable or even better predictive performance in classification  on publicly available data sets,
%namely on  schizophrenia, intractable seizures, Alzheimer and dementia patients.
%We postulate that this can be due the higher separability of the  EEG motifs found for these diseases compared to the EEG motifs for the MDD treatment. 
%However any direct comparison of the precision of our method on other tested data sets or  works  mentioned in the related work is to be taken with caution due to different  classification problems  or different sizes of the data sets, which are mostly  up to 30 patients.
%Previous  studies that analyze EEG signals in regards to depression, report precision on smaller datasets of up to 30 patients.
%\Mag{with  accuracy or F1 score  usually higher than 0.7. Thus it is necessary to see the accuracy and F1  by our framework on the MDD data set as a good accuracy, as  more data points introduce more  variance for the classifiers. Our data set of 176 patients is about six  times bigger than the size of common sets, thus  our precision needs to be interpreted  from this point of view when compared to these studies.}
%Moreover, our database includes a separate testing set, which allows the assessment of the final classifiers using an independent test set. While the works in the literature often employ cross-validation and apply it to   classification problems, they frequently lack evaluation on distinct test datasets. 
Accounting not only for the class imbalance but also for the gender imbalance proved to be crucial in achieving good performance in the case of MDD. In our proposed workflow we represent each patient group (class and gender) by the same number of motifs when computing the feature matrix. We have also considered the gender information in the experiments in the other psychiatric data sets where it was available, i.e., for the Alzheimer's disease and dementia. 

%\Blu{Have you done gender splitting also for the other 4 data sets? If not, we must write the discussion more carefully.} 
%Deciding on which motifs to include in the feature representation was done using the proposed motif ranking, by computing the \textit{difference score}, which describes how discriminative a motif is between the two classes.
% what else could have been done (if there was enough time), what can be i 

%\Mag{Can you shorten this paragraph?}

\vspace{0.1cm}

\noindent \textbf{Possible future work} The motif ranking computation, i.e., the difference score, represents a complex problem, given that it depends on several parameters. The z-normalized Euclidean distance, used by the difference score uses, is not appropriate for  comparison of similarity between pairs of time series of different lengths. In future  we want to  use  a length invariant score, as well as to explore the frequency bands gamma and delta and to investigate  longer motifs  that can also differentiate the two patient groups. 
Given that EEG channels are recorded simultaneously, one can consider motif discovery as a multi-dimensional motif search, having each electrode as a separate dimension. For this case,  other motif discovery algorithms can be selected.

%We conclude that to the best of our knowledge, our work is the first one applying motif discovery to extract features for prediction of   outcome of depression treatment and depression diagnostics in general. 
% more bands, 
% combine motifs from different bands
% usage of more powerful algorithms that do not need the parameter of the length... 
% try out multidimensional motif discovery, and explain how to apply it. 

% closing sentence.

%\pagebreak

%%%%%%%%%%%%%%%% ACM,  neni nase

\vspace{0.2cm}
\noindent \textbf{Acknowledgements}\\
This work was supported 
by the Austrian Science Foundation FWF (project I5113). 
%To Robert, for the bagels and explaining CMYK and color spaces.

%\newpage

\appendix
\newpage
\noindent {\Large \textbf{Appendix}}

\section{
More details to the EEG data sets}

\subsection{Preprocessing of the EEG Signal}
\label{preprocessing_eegs}

%\subsection{Preprocessing of the EEG signal}

The initial preprocessing of the EEG signals has been provided by the project members.
The preprocessing was done in MATLAB using the open-source toolbox EEGLAB \cite{delorme2004eeglab}. The following steps were performed:
\begin{itemize}
    \item  Downsample the signals with 1000 Hz sampling rate to 250 Hz, by keeping every 4th sample.
    \item Remove the first and last 30 seconds of the signal, as this period can contain a high number of artifacts.
    \item Use the Average Reference method \cite{averagereference1950eeg} to transform the EEG recordings. 

    (An EEG signal quantifies the electrical potential difference between the recording electrode and a reference electrode. In the Average Reference method, the reference electrode represents the average signal across all electrodes.)
    
	 \item  Apply a bandpass filter to keep only frequencies from 1 to 40 Hz.
	 \item Remove segments of a signal that contain high-power artifacts. 
  
  (The segments are determined by a window with no overlap and the ones containing problematic data are removed entirely. The length of the window is set to 2 seconds, as this ensures that every 2-second segment of the signal is continuous.)

  \item Since we are using each frequency band separately, we further downsample each signal for each band to its own Nyquist rate to save computation time. This means downsampling the theta, alpha, and beta bands to 16, 24, and 60 Hz respectively.
\end{itemize}

\begin{figure}[h!]
\includegraphics[width=11cm]{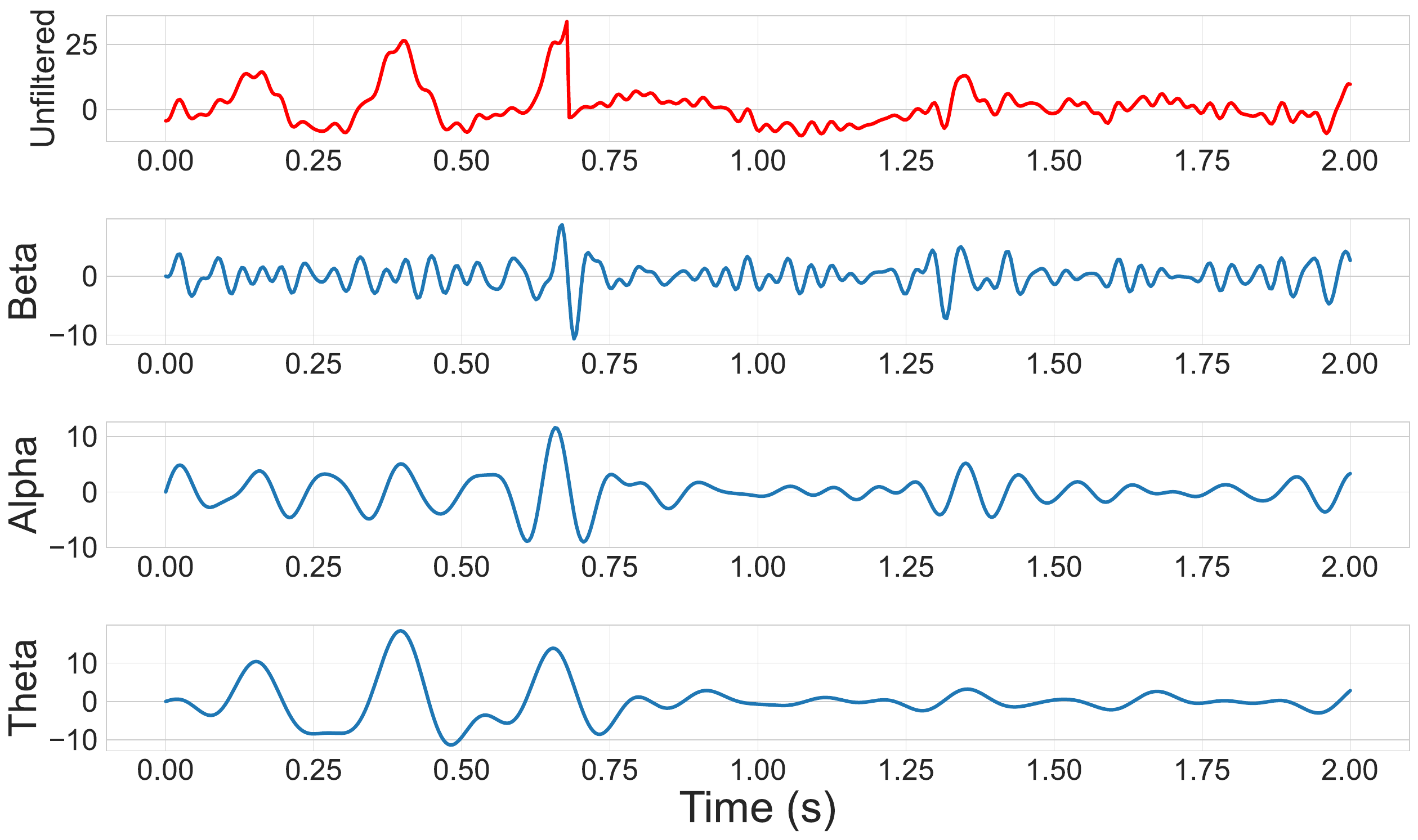}
\centering
\caption{Short segment of the EEG signal in channel Fp1 and the extracted frequency bands: beta, alpha and theta.}\label{frequency_bands}
\end{figure}

%Figure \ref{frequency_bands}  shows a short segment of electrode Fp1 of a patient across different frequency bands and the unfiltered (raw) signal. 

\subsection{Class and Gender Distribution}
\label{class_gender_distribution}

\begin{table}[h]
    \centering
    \begin{tabular}{l c c c}
    \hline
        & Male & Female & Total \\
        \hline
        Responders & 17 & 66 & 83 \\
        Non-responders & 31 & 62 & 93 \\
        Total & 48 & 128 & 176 \\
        \hline
    \end{tabular}
    \vspace{0.1cm}
    \caption{Class and gender distribution of the patients in the EEG database}
    %\Mag{Do the KDD template}}
    \label{training_and_test_sets}
\end{table}

\section{Details to the experiments}

\subsection{Feature separability}
\label{feature_matrix_projected_values}

\begin{figure}[h]
\centering
\textmd{Band: beta} \\
 %  \begin{minipage}{0.45\textwidth}
     \centering
     \includegraphics[width=.38\linewidth, scale=0.45]{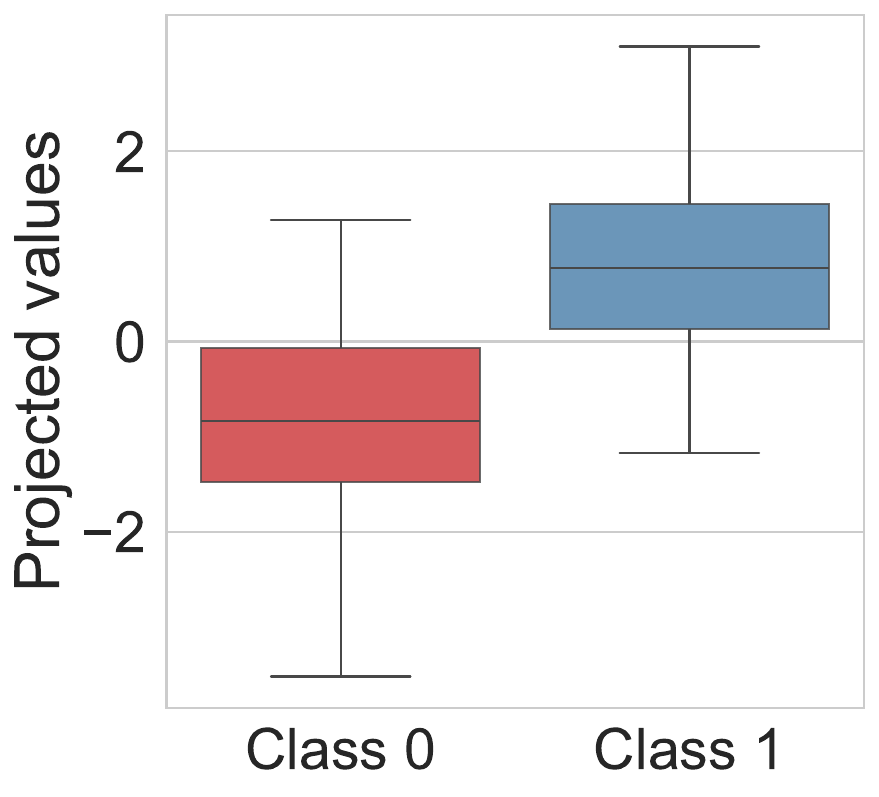}
%   \end{minipage}\hfill
%   \begin{minipage}{0.40\textwidth}
     \centering
     \includegraphics[width=.5\linewidth]{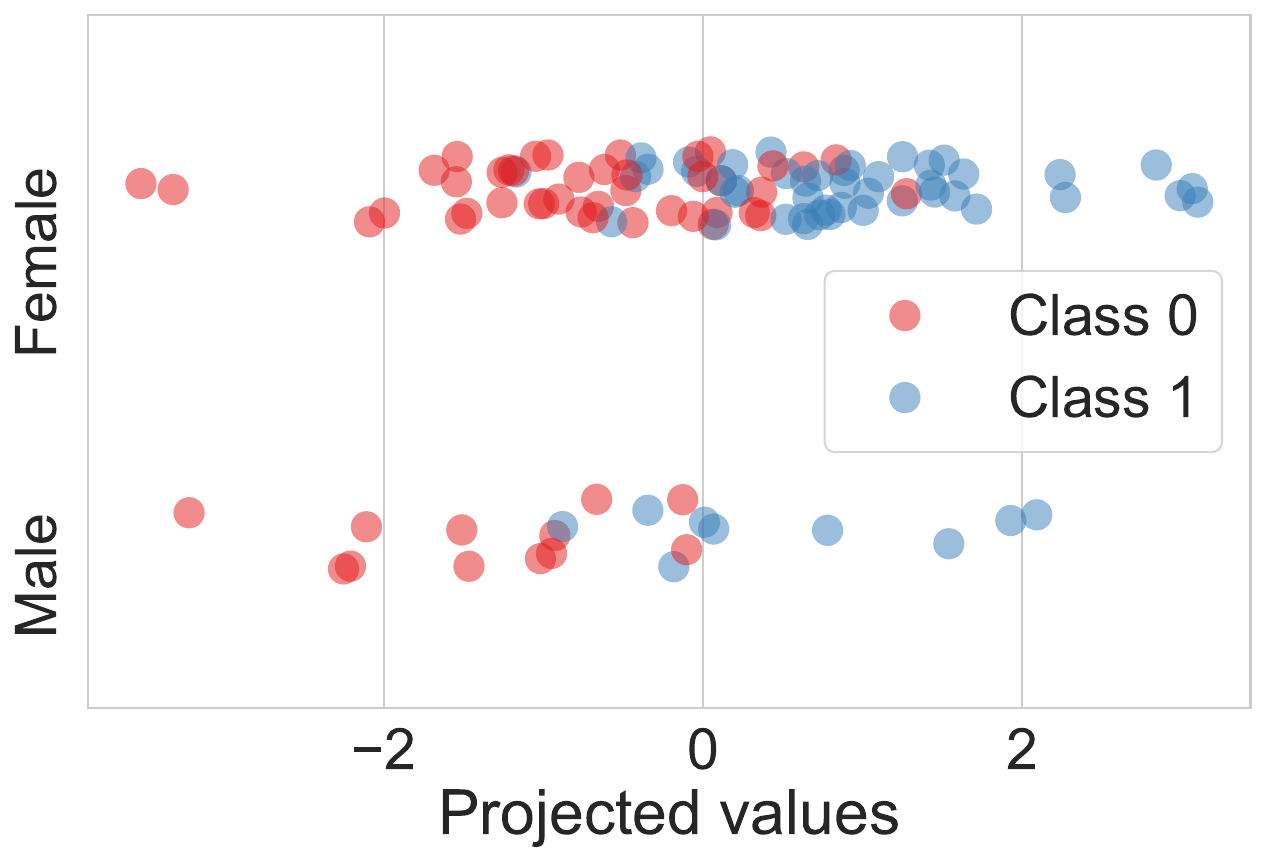}
%   \end{minipage} \\
\textmd{Band: theta} \\
%   \begin{minipage}{0.45\textwidth}
     \centering
     \includegraphics[width=.38\linewidth, scale=0.45]{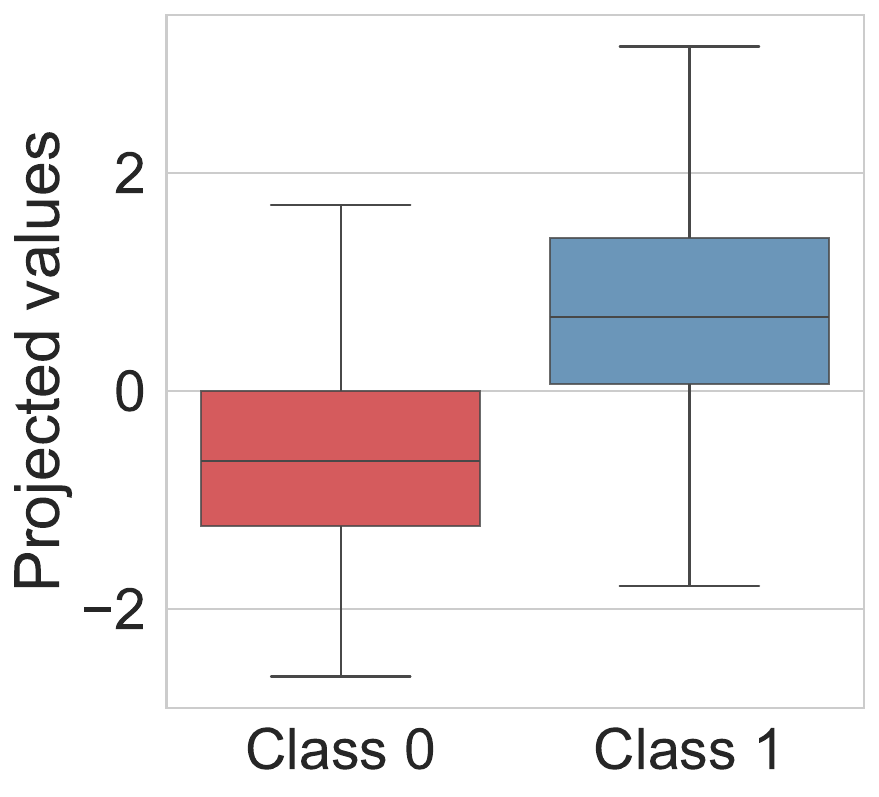}
  % \end{minipage}\hfill
  % \begin{minipage}{0.40\textwidth}
     \centering
     \includegraphics[width=0.5\linewidth]{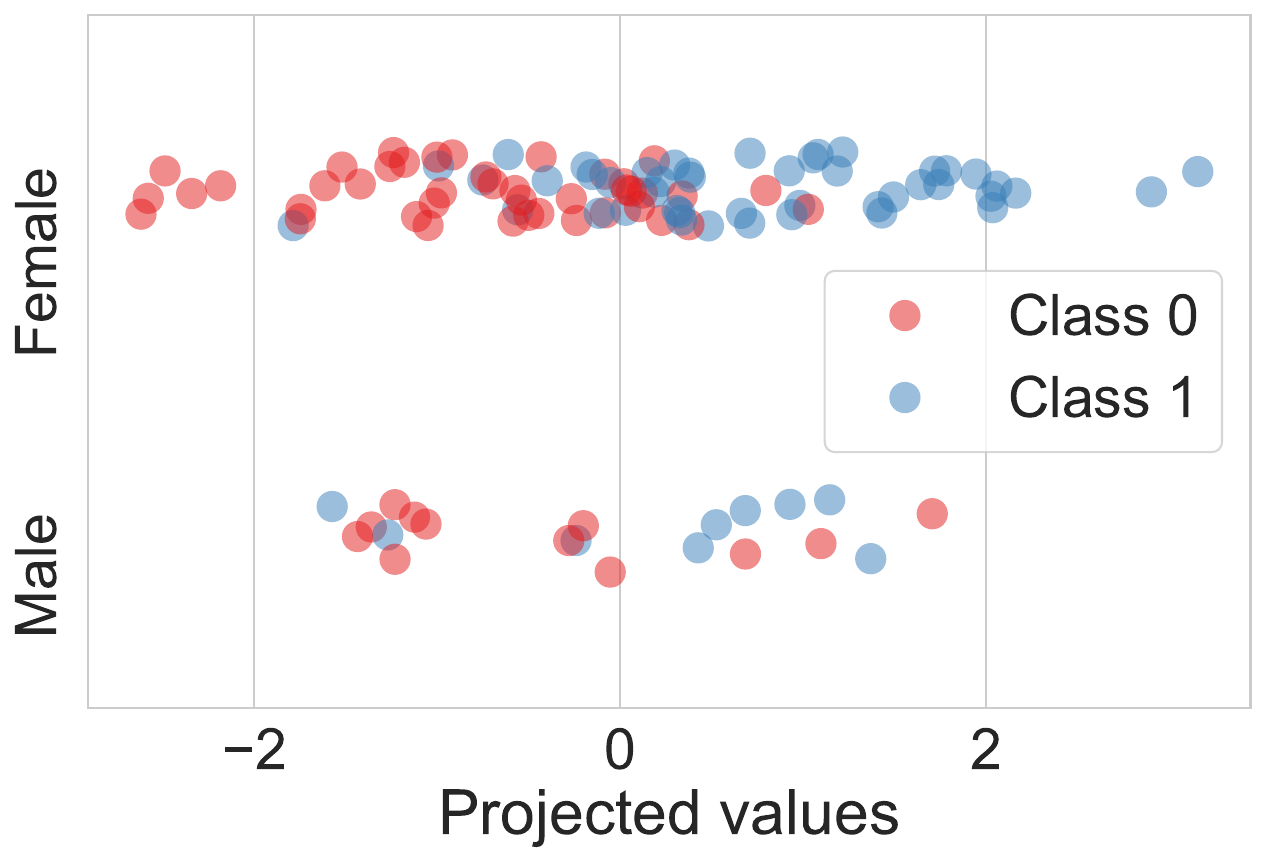}
  % \end{minipage}\\
    \caption{Examples of the distribution of the projected values of the feature matrix of the beta and theta band.}

\end{figure}

\subsection{Hyperparameter tuning}
\label{hyperparameters}

For choosing the most optimal hyperparameters for the model, a 5-fold cross-validation is used. Each model had the following hyperparameters that were tuned:
\begin{itemize}
    \item SVM
    \begin{itemize}
        \item kernel: ["linear", "rbf"]
        \item C: [0.0005, 0.001, 0.01, 0.1, 0.5]
        \item penalty: ["l1", "l2"]
    \end{itemize}
    \item Decision Tree 
    \begin{itemize}
        \item criterion: ["gini", "log\_loss"]
        \item max\_depth: [3, 4, 5, 10, 20]
    \end{itemize}
    \item Random Forest
    \begin{itemize}
        \item n\_estimators: [5, 10, 15, 20]
        \item max\_depth: [3, 4, 5, 10, 20]
        \item min\_samples\_leaf: [1, 2, 3, 4]
        \item min\_samples\_split: [2, 3, 4]
    \end{itemize}
    \item Logistic Regression
    \begin{itemize}
        \item C: [0.1, 0.5, 0.7, 1]
        \item penalty: ["l1", "l2"]
    \end{itemize}
    \item Multi-layer perception
        \begin{itemize}
            \item alpha: [0.01, 0.05, 0.1, 0.5]
            \item hidden\_layer\_sizes: [(5, 10, 5), (10, 10), (20, 10, 20), (20, 20)]
        \end{itemize}
\end{itemize}

\subsection{Cross Validation}
\label{cross_val_accuracy}

\begin{figure}[!htb]
\centering
\textmd{Cross Validation}\\
\begin{minipage}{0.9\textwidth}
     \centering
     \includegraphics[width=0.49\linewidth]{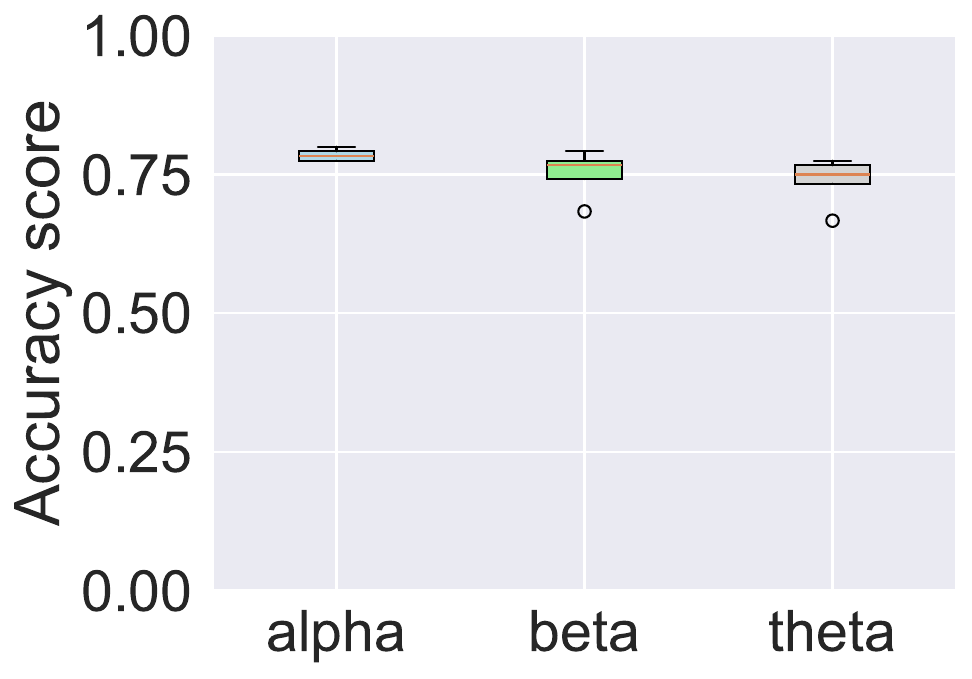}
 %  \end{minipage}\hfill
 % \begin{minipage}{0.47\textwidth}
     \centering
     \includegraphics[width=0.49\linewidth]{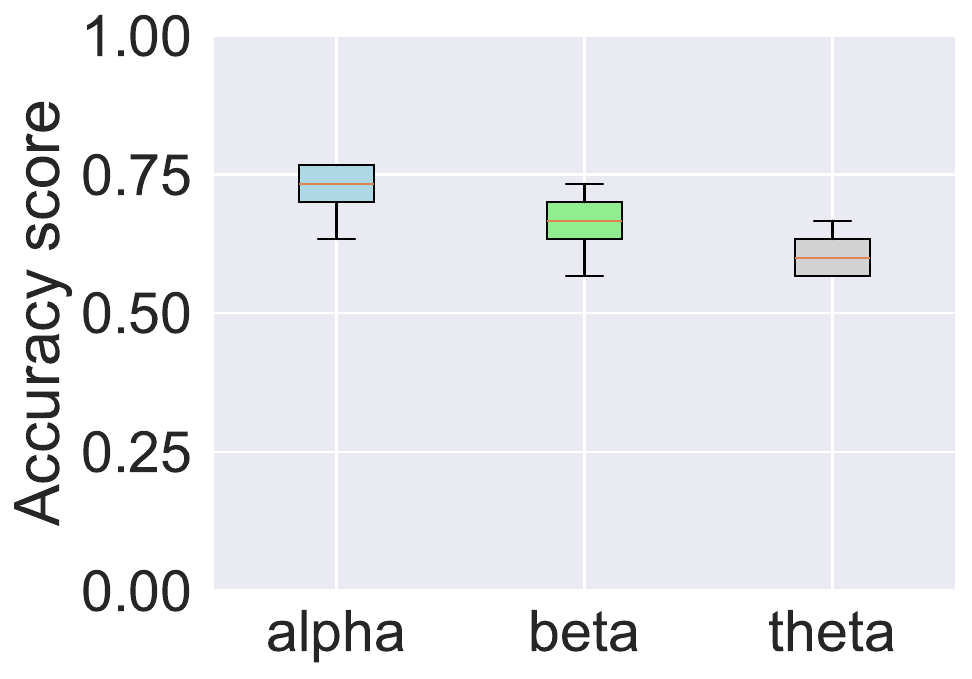}
   \end{minipage} 
\caption{Mean Accuracy scores of training (left) and validation (right) for all three frequency bands of the best performing classifier.}\label{val_boxplot_acc}
\end{figure}
\newpage
\subsection{Precision on other psychiatric data sets}
\label{other_psychiatric_datasets}

The results can be found in Table~\ref{otherdata sets}.

\begin{table}[h]
\centering
  \caption{The best evaluation results on other psychiatric data sets within each band.}\label{psy_datasets_results_bands}
  \begin{tabular}{lccccc}
  \\
  \hline
     Dataset & Band & \multicolumn{2}{c}{Training} & \multicolumn{2}{c}{Validation}  \\
   % \cline{3-6}
    & & Accuracy & F1 & Accuracy & F1  \\
     \hline
     \multirow{3}{*}{Schizophrenia} 
     & alpha & 0.927 &	0.931 &	0.82 &	0.821 \\
     & beta &  0.898 &	0.907 &	0.857 &	0.877\\
    & theta & 0.767 & 0.799 & 0.762 & 0.789 \\
    \multirow{3}{*}{Seizures} & alpha & 0.875 & 0.875 & 0.675 & 0.617\\
        & beta & 0.711 & 0.749 & 0.624 & 0.681 \\
        & theta & 0.834 & 0.831	& 0.787	& 0.784  \\
    \multirow{3}{*}{Alzheimer's} 
        & alpha & 0.896 &  0.88 & 0.8 &  0.774\\
        & beta & 0.9 & 0.886 & 0.861 &  0.845 \\
        & theta & 0.868 &	0.847 &	0.845	& 0.823 \\
    \multirow{3}{*}{Dementia} %&
        & alpha  &  0.797 & 0.827 & 0.808 & 0.835 \\
        & beta   & 0.971 & 0.974 & 0.807 & 0.82 \\
        & theta  & 0.954 &	0.951&	0.87	& 0.865 \\
  \hline
\end{tabular}\label{otherdata sets}
\end{table}

\end{document}